%% file: iclr2024_conference.tex
\documentclass{article} 
\usepackage{iclr2024_conference,times}

\input{math_commands.tex}

\usepackage{hyperref}
\usepackage{url}

\usepackage{graphicx}
\usepackage{algorithm}
\usepackage{algorithmic}

\usepackage{booktabs}
\usepackage{float}
\usepackage{amsmath}
\usepackage{amsfonts,amssymb}
\usepackage{mathrsfs}
\usepackage{multirow}
\usepackage{makecell}
\usepackage{newfloat}
\usepackage{listings}
\usepackage{enumitem}
\usepackage{tabularx}
\usepackage{array}
\usepackage[hypcap=false]{caption}
\usepackage{nicefrac}       
\usepackage{microtype}      
\usepackage{xcolor}         
\usepackage{comment}
\usepackage{mathtools}
\usepackage{subcaption}
\usepackage{tablefootnote}
\usepackage{textcomp} 
\usepackage{scalerel}
\usepackage{wrapfig}
\usepackage{colortbl}
\usepackage{longtable}
\usepackage{CJKutf8}
\title{TRACE: A Comprehensive Benchmark for Continual Learning in Large Language Models}

\author{
    {\normalsize
     \textbf{Xiao Wang}$^{\bigstar}$\thanks{Equal contribution}\quad
     \textbf{Yuansen Zhang}$^{\bigstar*}$\quad
     \textbf{Tianze Chen}$^{\bigstar*}$\quad
     }\\
    {\normalsize
    \textbf{Songyang Gao}$^{\bigstar}$\quad 
    \textbf{Senjie Jin}$^{\bigstar}$\quad
    \textbf{Xianjun Yang}$^{\clubsuit}$\quad
    \textbf{Zhiheng Xi}$^{\bigstar}$\quad
    }\\
    {\normalsize
    \textbf{Rui Zheng}$^{\bigstar}$\quad
    \textbf{Yicheng Zou}$^{\spadesuit}$\quad
    \textbf{Tao Gui}$^{\blacklozenge}$\thanks{Corresponding Author}\quad
    \textbf{Qi Zhang}$^{\bigstar \dagger}$\quad
    \textbf{Xuanjing Huang}$^{\bigstar}$
    }\\
  {$^\bigstar$ \normalsize School of Computer Science, Fudan University, Shanghai, China} \\
  {$^\blacklozenge$ \normalsize Institute of Modern Languages and Linguistics, Fudan University, Shanghai, China} \\
  {$^\clubsuit$ \normalsize University of California, Santa Barbara} \\
  {$^\spadesuit$ \normalsize Shanghai AI Laboratory}\\
  \texttt{\normalsize \{xiao\_wang20,qz,tgui\}@fudan.edu.cn}
}

\iclrfinalcopy 
\begin{document}

\maketitle

\begin{abstract}

Aligned large language models (LLMs) demonstrate exceptional capabilities in task-solving, following instructions, and ensuring safety. However, the continual learning aspect of these aligned LLMs has been largely overlooked.
Existing continual learning benchmarks lack sufficient challenge for leading aligned LLMs, owing to both their simplicity and the models' potential exposure during instruction tuning.
In this paper, we introduce TRACE, a novel benchmark designed to evaluate continual learning in LLMs. TRACE consists of 8 distinct datasets spanning challenging tasks including domain-specific tasks, multilingual capabilities, code generation, and mathematical reasoning. 
All datasets are standardized into a unified format, allowing for effortless automatic evaluation of LLMs.
Our experiments show that after training on TRACE, aligned LLMs exhibit significant declines in both general ability and instruction-following capabilities.
For example, the accuracy of llama2-chat 13B on gsm8k dataset declined precipitously from 28.8\% to 2\% after training on our datasets.
This highlights the challenge of finding a suitable tradeoff between achieving performance on specific tasks while preserving the original prowess of LLMs.
Empirical findings suggest that tasks inherently equipped with reasoning paths contribute significantly to preserving certain capabilities of LLMs against potential declines. 
Motivated by this, we introduce the Reasoning-augmented Continual Learning (RCL) approach. RCL integrates task-specific cues with meta-rationales, effectively reducing catastrophic forgetting in LLMs while expediting convergence on novel tasks.

\end{abstract}

\section{Introduction}
Large Language Models (LLMs) \citep{OpenAI2023GPT4TR, touvron2023llama} have revolutionized natural language processing through a two-step process: initial pretraining on extensive corpora, followed by fine-tuning on human-generated instructions and preference data, aligning them with human language and intentions.
Aligned LLMs have showcased impressive capabilities and ensured safer responses. However, as the demands for language models grow, there's a pressing need to enhance their abilities in areas such as domain-specific knowledge \citep{wu2023bloomberggpt, li2023starcoder}, multilingual proficiency \citep{huang2023c}, complex task-solving \citep{surameery2023use}, and tool usage \citep{Qin2023ToolLLMFL}.
Yet, retraining and realigning them from scratch to meet these demands is impractical due to prohibitive training costs and the challenge of acquiring high-quality data.
Therefore, incrementally training existing Aligned LLMs through continual learning (CL \citep{Wang2023ACS}) is crucial. 
This prompts the pressing question: \textit{To what degree do Aligned LLMs exhibit catastrophic forgetting when subjected to incremental training?}

Existing continual learning benchmarks \citep{Zhang2015CharacterlevelCN, Scialom2022FinetunedLM,razdaibiedina2023progressive} are not suitable for evaluating the state-of-the-art LLMs. Firstly, many of these benchmarks predominantly consist of simplistic natural language understanding datasets. These tasks, due to their inherent simplicity, fail to challenge the capabilities of large-scale models adequately. 
Furthermore, a significant drawback lies in the fact that some of these datasets have previously appeared in the instruction tuning \citep{Chung2022ScalingIL} sets of these LLMs, suggesting that the models may have already learned these tasks during training.
Secondly, prior benchmarks have primarily focused on metrics that assess the performance of the models on target sequential tasks. 
Yet, for aligned models, aspects like generalization to new tasks, the ability to follow human instructions, and safety preservation are of paramount importance. Regrettably, these dimensions have not been extensively studied or incorporated into assessments.

To facilitate further research, we present \textbf{TRACE}, a continual learning benchmark designed for aligned Large Language Models.
Our benchmark consists of 8 distinct datasets spanning challenging tasks including domain-specific tasks, multilingual capabilities, code generation, and mathematical reasoning. 
To ensure task balance, we have sampled 5,000 instances for each task, and for classification tasks, we have ensured an equal distribution of samples across all classes.
Additionally, all datasets have been standardized into a unified format, simplifying the evaluation process.
To evaluate continual learning in aligned LLMs, we introduce three metrics: "General Ability Delta," "Instruction Following Delta," and "Safety Delta" to assess models' forgetfulness in such scenarios.

We conduct a comprehensive evaluation of 5 aligned LLMs on TRACE. 
Evaluation results reveal several key findings: 1) Nearly all models exhibit a significant decline in general abilities, especially in math and reasoning. For instance, when trained on TRACE, the accuracy of llama2-chat 13B on the gsm8k dataset dropped from 28.8\% to a mere 2\%.
2) Unlike other skills, LLMs' multilingual abilities generally improve. For example, llama2-chat 7B's performance on the TydiQA dataset surged from an F1 score of 23.47 to 33.23.
3) Full-parameter training, compared to LoRA training, more easily fits the target tasks, but it also leads to a more pronounced decline in general abilities.
4) LLMs' instruction-following capabilities also suffer a significant reduction after continual learning.

\begin{figure*}[t] \vspace{-0.5cm}
\centering    \includegraphics[width=1\textwidth]{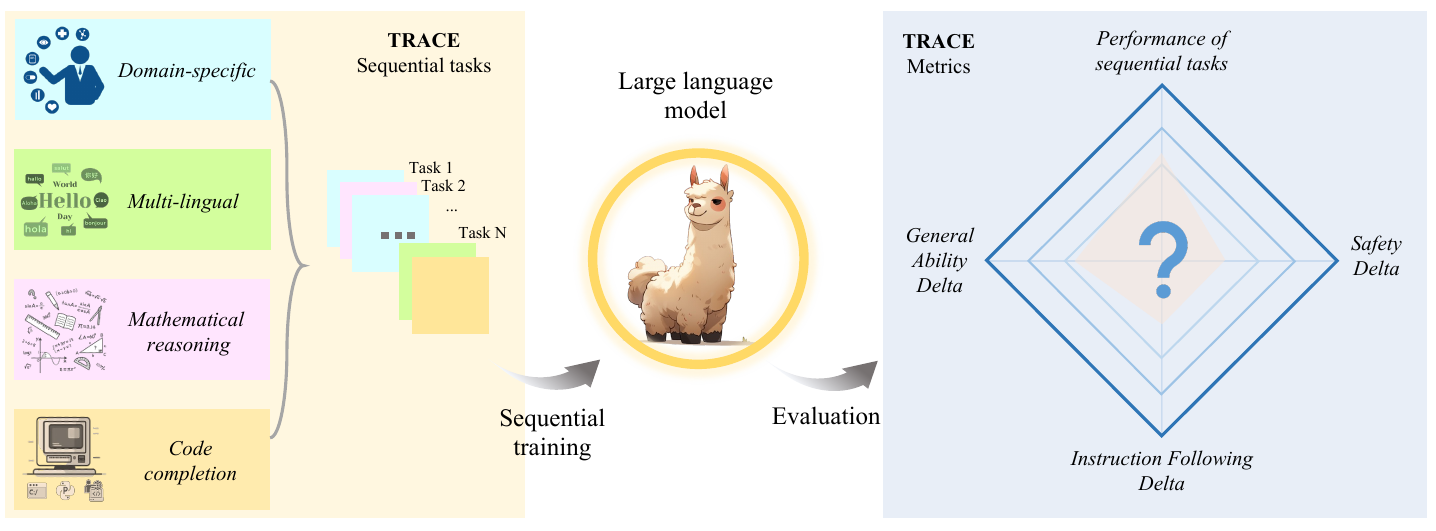} 
    \caption{An overview of TRACE benchmark. TRACE consists of two main components: 1) A selection of eight datasets constituting a tailored set of tasks for continual learning, covering challenges in domain-specific tasks, multilingual capabilities, code generation, and mathematical reasoning. 2) A post-training evaluation of LLM capabilities. In addition to traditional continual learning metrics, we introduce General Ability Delta, Instruction Following Delta, and Safety Delta to evaluate shifts in LLM's inherent abilities.}\label{fig:TRACE} 
\end{figure*}

Through experimentation, we observed that tasks augmented with reasoning paths are notably effective in preserving certain capabilities of LLMs, preventing them from substantial declines. Such findings lead us to ponder on leveraging a model's inherent strengths for rapid transfer on new tasks, rather than starting the learning curve from scratch. This motivation birthed our novel training strategy: Reasoning-augmented Continual Learning (RCL). RCL prompts the model to generate task analyses and rationales during training. As our results indicate, this approach not only boosts performance on target tasks but also significantly upholds the inherent strengths of LLMs. \footnote{The dataset, code can be found at \url{https://github.com/BeyonderXX/TRACE}}.

\section{Related Work}
\label{Related Work}
\subsection{Continual Learning}
Continual learning \citep{Wang2023ACS} aims to develop learning algorithms that can accumulate knowledge on non-stationary data. Existing works can be broadly categorized into rehearsal-based, regularization-based, and architecture-based approaches. Rehearsal-based approaches \citep{lopez2017gradient, de2019episodic} leverage a memory buffer that stores examples from previous tasks, training the model jointly with the current task. 
Experience replay (ER) \citep{rolnick2019experience} is a common strategy employed in rehearsal-based approaches and serves as a strong baseline. 
Regularization-based approaches \citep{kirkpatrick2017overcoming, smith2023continual} incorporate additional terms into the loss function to penalize changes in crucial weights. 
For example, Orthogonal Gradient Descent (OGD) \citep{farajtabar2020orthogonal} constrains the parameters to move within the orthogonal space defined by the gradients of previous tasks, Elastic Weight Consolidation (EWC) \citep{kirkpatrick2017overcoming} constraint important parameters to stay close to their previous ones through, Gradient Episodic Memory (GEM) \cite{lopez2017gradient} leverages episode memories to avoid forgetting. 
Architecture-based approaches \citep{Wang2023EIP, razdaibiedina2023progressive} focus on dynamically expanding model capacity or isolating existing model weights to mitigate interference between new and old tasks. 
Progressive Prompts \citep{razdaibiedina2023progressive} learns separate prompts for each incoming task and sequentially concatenates them with previously learned prompts. 

\subsection{CL Benchmarks in NLP}
The most recognized CL benchmark for NLP encompasses five text classification datasets introduced by \citet{Zhang2015CharacterlevelCN}, including AG News, Amazon Reviews, Yelp Reviews, DBpedia, and Yahoo Answers. Building upon this, \citet{razdaibiedina2023progressive} proposed a long CL benchmark which fuses the aforementioned five datasets with an additional four from the GLUE benchmark \citep{wang2018glue}, five from the SuperGLUE benchmark \citep{wang2019superglue}, and the IMDB dataset \citep{Maas2011LearningWV}.
While these datasets have been incorporated into the Flan collection \citep{Chung2022ScalingIL} and are widely used for current SoTA LLMs, their ubiquity has rendered them less suitable as CL benchmarks for LLMs.
Taking a different approach, \citep{Scialom2022FinetunedLM} focuses on English language generation tasks. In a subsequent study, \cite{Luo2023AnES} conducted an analysis of catastrophic forgetting on Bloomz \citep{Scao2022BLOOMA1} using \cite{Scialom2022FinetunedLM}'s datasets. However, this benchmark is limited in scope as it solely emphasizes Natural Language Generation (NLG) tasks and is restricted to English, thus lacking task diversity.
In contrast, TRACE offers a diverse and challenging array of sequential tasks. Additionally, it evaluates aligned models on aspects such as general capability, adherence to instructions, and shifts in safety measures.

\subsection{Chain-of-Thought}
LLMs have shown advanced step-by-step reasoning capabilities, known as Chain-of-Thought (CoT) reasoning \citep{wei2022chain}.
Zero-shot-CoT \citep{kojima2022large} illustrates the profound impact of simply prefacing a reasoning sequence with the sentence, "Let’s think step by step." Least-to-most \citep{zhou2022least}, actively prompts LLMs to segment complex problems into smaller tasks. ScienceQA \citep{lu2022learn} highlights the efficacy of CoT in LLMs, particularly beneficial for filling the void in datasets within the scientific realm.
Fine-tune-CoT \citep{ho2022large} exploits the prowess of extensive LLMs to generate reasoning samples, further fine-tuning smaller models. 
Despite these advancements, the application of CoT in continual learning remains unexplored. Our benchmark, TRACE, showcases that generating explanations can not only accelerate the learning process but also significantly mitigate the forgetting in their foundational capabilities.

\section{Preliminaries}
\label{Preliminaries}
Continual learning \citep{Ke2022ContinualLO, Wang2023ACS} focuses on developing learning algorithms to accumulate knowledge on non-stationary data. In supervised continual learning, a sequence of tasks $\left\{\mathcal{D}_1, \ldots, \mathcal{D}_T\right\}$ arrive in a streaming fashion. Each task $\mathcal{D}_t=\left\{\left(\boldsymbol{x}_i^t, y_i^t\right)\right\}_{i=1}^{n_t}$ contains a separate target dataset, where $\boldsymbol{x}_i^t\in \mathcal{X}_t$ , $\boldsymbol{y}_i^t\in \mathcal{Y}_t$. 
A single model needs to adapt to them sequentially, with only access to $\mathcal{D}_t$ at the t-th task. 
In general, given a prediction model $h_{\Theta}$ parameterized by $\Theta$, continual learning seeks to optimize for the following objective across all tasks:
\begin{equation}
\max _{\Theta} \sum_{k=1}^T \sum_{x, y \in \mathcal{D}_k} \log p_{\Theta}(y \mid x)
\end{equation}

In this paper, we utilize overall performance (OP \citep{chaudhry2018riemannian}) and backward transfer (BWT \citep{lopez2017gradient}) scores as the main metrics. After incrementally learning the t-th task, the model's score on the i-th task (where $i \geq t$) is denoted as $ R_{t,i}^D $. The overall performance and backward transfer score are calculated using the following formulas:
\begin{equation}
    OP_t = \frac{1}{t} \sum_{i=1}^t R_{t, i}^D
\end{equation}

\begin{equation}
    BWT_t = \frac{1}{t} \sum_{i=1}^{t-1}\left(R_{t, i}^D - R_{i, i}^D\right)
\end{equation}

\section{TRACE: A Comprehensive Benchmark for CL in LLMs}
\label{trace method}
TRACE is designed to offer a comprehensive continual learning evaluation for LLMs. Illustrated in Figure \ref{fig:TRACE}, TRACE encompasses two primary components: a curated set of tasks tailored for continual learning, followed by an in-depth evaluation of an LLM's post-training capabilities. 
In this section, we detail TRACE's sequential tasks and introduce our evaluation metrics. In Section \ref{main results}, we evaluate five models using TRACE and present our key findings.

\subsection{Data Creation}
There are three principles for the creation of TRACE. First, the datasets should be novel enough that most LLMs have not been trained on them. Second, they should be challenging for large language models. Third, a variety of tasks should be covered in our benchmark. 

According to these three principles, in this section, we will provide a detailed introduction to the data collection process for each dataset. As these datasets have varying sizes, we create a balanced version by randomly sampling 5000 training examples and 2000 testing examples from the original datasets. As shown in Table \ref{Dataset statis}, we get 40,000 training examples and 16,000 testing examples in total. 

\textbf{Domain-Specific.} These tasks require specific knowledge, so models may perform poorly if they have not appeared frequently enough in the training data. We select three datasets, ScienceQA \citep{lu2022learn}, FOMC \citep{shah2023trillion} and MeetingBank \citep{hu2023meetingbank}. 
ScienceQA is a multi-hop QA dataset collected from elementary and high school science curricula, with a rich domain diversity from natural science, social science, and language science, requiring the model of reasoning ability and science knowledge. 
As we only test the performance of language models, only the examples without multi-modal contexts are included in TRACE. FOMC is a hawkish-dovish classification task, which is novel in the financial domain. The dataset is divided into three subsets: data on meeting minutes, press conference data, and speech data. We use a combination of them. MeetingBank is a new benchmark dataset for city council meeting summarization, an unstudied domain. It demands a global understanding of the whole long context.

\textbf{Multi-lingual.} 
The cross-lingual ability of large language models is limited due to vocabulary and pre-training corpus. For instance, LLaMA's vocabulary contains few Chinese tokens, affecting its efficiency with Chinese text. \citep{cui2023efficient} expand LLaMA's Chinese vocabulary and fine-tune with additional Chinese corpus. Yet, capabilities for other languages can be forgotten after training in a specific language, making it vital to evaluate cross-lingual ability in our benchmark.
We select C-STANCE\citep{zhao-etal-2023-c} and 20Minuten\citep{rios-etal-2021-new} as multi-lingual datasets. C-STANCE is the first Chinese dataset for zero-shot stance detection collected from Sina Weibo, one of the most popular Chinese social media sites. It includes two challenging subtasks: target-based stance detection and domain-based stance detection. In TRACE, we include the target-based one, which means the targets in testing examples are unseen during training. 20Minuten is a text simplification dataset consisting of full articles paired with shortened, simplified summaries from the Swiss news magazine. We use this dataset to evaluate the ability to generate German text.

\textbf{Code completion.} Code completion is another challenging task to evaluate long context modeling ability\citep{bai2023longbench}, and it is one of the most widely used features in software development through IDEs. We select the line-level code completion task of CodeXGLUE\citep{lu2021codexglue}, which requires the model to generate the next line given the lengthy code input. The corpus Py150 contains 150,000 Python programs collected from GitHub repositories. Since the golden labels of the testing dataset are not available by \cite{lu2021codexglue}, we randomly divide each Python code in Py150 into two parts, taking the first part as inputs and the next line as labels.

\textbf{Mathematical reasoning.} Mathematical problems are always used to evaluate the reasoning ability of models. NumGLUE\citep{mishra2022numglue} is an 8-task benchmark far from solved including state-of-the-art large-scale language models performing significantly worse than humans. We include the first two tasks of NumGLUE because they are freshly collected data and intelligent modifications of already existing datasets. Both of the two tasks require arithmetic reasoning ability. 
It is worth noting that both datasets have original labels consisting only of numbers, without associated inference processes.

\subsection{CL Metrics Design}
\label{CL metrics}
Unlike traditional continual learning benchmarks focused on sequential target tasks, evaluating aligned LLMs should also account for the preservation of their inherent capabilities. SoTA LLMs, through instruction tuning, exhibit impressive task-solving abilities. Aligning these models with human preferences further boosts their safety and usefulness. Hence, TRACE introduces a broader evaluation, including three unique metrics: "General Ability Delta," "Instruction Following Delta," and "Safety Delta." These measure the changes in LLM's general capability, instruction adherence, and response safety post-continual learning.

\textbf{General Ability Delta} is designed to assess the change in performance of an LLM on generic tasks after training on sequential target tasks.
Let's consider a set of general tasks denoted as $\{G_1, ... , G_M\}$. The baseline performance of the initial LLM on the i-th general task is represented by $ R_{0,i}^G $. After incrementally learning up to the t-th task, the score on the i-th general task becomes $ R_{t,i}^G $. The "General Ability Delta" after training on the t-th task, represented as $ \Delta R_t^G $, is given by:
\begin{equation}
\Delta R_t^G = \frac{1}{M} \sum_{i=1}^M ( R_{t, i}^G - R_{0, i}^G)
\end{equation}

\textbf{Instruction Following Delta} measures the change in a model's ability to follow instructions after training on sequential tasks. Using a set of datasets, represented as ${I_1, ..., I_N}$, the initial LLM performance on the i-th task is $ R_{0,i}^I $. After incremental learning to the t-th task, its score on the i-th task is $ R_{t,i}^I $. The change, represented by $ \Delta R_t^I $, is computed as:
\begin{equation}
\Delta R_t^I = \frac{1}{N} \sum_{i=1}^N (R_{t, i}^I - R_{0, i}^I)
\end{equation}

\textbf{Safety Delta} quantifies the change in a model's response safety after sequential training. Using a set of datasets designed for safety evaluation, denoted as ${S_1, ..., S_L}$, the initial safety metric on the i-th dataset is $ R_{0,i}^S $. After training up to the t-th task, its score on the i-th dataset is $ R_{t,i}^S $. The change after the t-th task, represented by $ \Delta R_t^S $, is computed as:
\begin{equation}
\Delta R_t^S = \frac{1}{L} \sum_{i=1}^L (R_{t, i}^S - R_{0, i}^S)
\end{equation}

\subsection{Experimental Setup}
\subsubsection{Baselines}
We evaluate the performance of LLMs in a continual learning setting using four approaches—three requiring training and one not:

\textbf{Sequential Full-Parameter Fine-Tuning (SeqFT)}: This method involves training all model parameters in sequence. 

\textbf{LoRA-based Sequential Fine-Tuning (LoraSeqFT)}: Only the low-rank LoRA matrices are fine-tuned, leaving the LLM backbone fixed \citep{Hu2021LoRALA}. This method is chosen based on prior findings of reduced forgetting with "Efficient Tuning" \citep{Liu2023PromptLT}.

\textbf{Replay-based Sequential Fine-Tuning (Replay)}: Replay, a common continual learning strategy, is employed for its simplicity and effectiveness. We incorporate alignment data from LIMA into the replay memory, replaying 10\% of historical data.

\textbf{In-Context Learning (ICL)}: Task demonstrations are supplied as part of the language prompt, acting as a form of prompt engineering \citep{Brown2020LanguageMA}. A 6-shot setting is used for our experiments.

To evaluate the resilience of safety alignment models from diverse training backgrounds and strategies, we select five aligned models from three organizations: Meta: LLaMa-2-7B-Chat, LLaMa-2-13B-Chat \citep{touvron2023llama}, BaiChuan: Baichuan 2-7B-Chat \citep{baichuan2023baichuan2}, and Large Model Systems Organization: Vicuna-13B-V1.5, Vicuna-7B-V1.5 \citep{vicuna2023}.

\subsubsection{Datasets}
To evaluate a model's \textit{general ability}, we assess across five key dimensions: Factual Knowledge, General Reasoning, Multilinguality, Commonsense Reasoning, and Reading Comprehension. 

\textbf{Factual Knowledge}: We use the Massive Multitask Language Understanding dataset (MMLU \citep{hendrycks2020measuring}) with questions on 57 subjects, from elementary to professional levels. Following LLaMa-2 \citep{touvron2023llama}, we report 5-shot accuracy based on answer perplexity.

\textbf{General Reasoning}: Evaluated using Big-Bench-Hard (BBH \citep{suzgun2022challenging}) with 23 tasks from Big-Bench \citep{Ghazal2013BigBenchTA}. The evaluation uses chain-of-thought prompts with 3-shot in-context examples, and EM scores are reported.

\textbf{Multilinguality}: We use TyDiQA \citep{clark2020tydi}, a multilingual QA benchmark with 11 languages. Using the gold-passage setup, 0-shot F1 scores are reported.

\textbf{Commonsense Reasoning}: Evaluated using PIQA \citep{bisk2020piqa}. Following LLaMa-2 \citep{touvron2023llama}, we report 0-shot accuracy based on answer perplexity.

\textbf{Reading Comprehension}: Assessed with BoolQ \citep{clark2019boolq}, containing 15942 questions. We report 0-shot accuracy based on answer perplexity.

For \textit{instruction-following capability}, we use Self-instruct dataset \citep{wang2022self}, which is user-oriented and comprises a diverse set of 175 prompts, encompassing areas like email writing, social media, productivity tools, entertainment, and programming. We also employ LIMA dataset \citep{zhou2023lima} LIMA assembles 300 prompts, primarily sourced from community Q\&A forums and supplemented by manually authored examples.

For assessing changes in \textit{safety}, we leverage the CoNa dataset \citep{bianchi2023safety}. This corpus encompasses 178 expert-annotated samples, specifically curated to address instructions associated with hateful speech generation.

\begin{table}[b] \vspace{-0.3cm}
    \centering
    \caption{OP(BWT) for all the baseline models and 3 baseline methods.}
    \begin{tabular}{c|c|c|c|c}
    \hline
        & ICL&SeqFT & LoraSeqFT & Replay \\
        \hline
        LLaMA-2-7B-Chat &$38.9$& $48.7(-8.3\%)$ & $12.7(-45.7\%)$ & $55.5(2.6\%)$ \\
        LLaMA-2-13B-Chat &$41.9$& $49.9(-7.0\%)$ & $28.0(-36.5\%)$ & $56.6(0.4\%)$\\
        Vicuna-7B-V1.5 &$42.2$& $49.2(-8.4\%)$ & $33.4(-23.7\%)$ & $55.3(0.2\%)$ \\
        Vicuna-13B-V1.5 &$46.9$& $51.7(-5.9\%)$ & $31.6(-28.4\%)$ & $56.9(0.6\%)$ \\
        Baichuan2-7B-Instruct &$44.6$& $43.4(-15.4\%)$ & $43.8(-9.0\%)$ & $51.7(1.1\%)$ \\
        \hline
    \end{tabular}

    \label{tab:CL results}
\end{table}

\subsubsection{Metrics}
As mentioned in Section \ref{Preliminaries}, we measure the performance of LLMs in continual learning tasks using Overall Performance and Backward Transfer Score. To check how well LLMs keep their original abilities, we use General Ability Delta, Instruction Following Delta, and Safety Delta. More details are in Section \ref{CL metrics}. For evaluating instruction-following and safety, we score with GPT-4 \citep{OpenAI2023GPT4TR}. More details about this scoring can be found in the Appendix \ref{gpt4eval}.

\subsubsection{Implementation Details}
The detailed settings can be found in Appendix \ref{sec:imple details}.

\subsection{Main Results}
\label{main results}

\subsubsection{Performance of Target Sequential Tasks}
Table \ref{tab:CL results} showcases the performance of five distinct LLMs on TRACE benchmark, after their continual learning phase. From this evaluation, we can draw the following conclusions:

\textbf{In-Context Learning (ICL) Performance}: ICL methods generally perform lower than SeqFT and Replay methods. This suggests that the TRACE benchmark is indeed challenging, and LLMs can't readily identify solutions just through simple demonstrations.

\textbf{Replay Performance}: Among all the baselines, Replay achieved the highest OP score. With its BWT score being positive, it indicates that Replay effectively retains its performance on sequential tasks without significant forgetting. This makes Replay a straightforward and efficient strategy in a continual learning context.

\textbf{Full Parameter Training vs. LoRA}: Full parameter training demonstrates better task-specific adaptability compared to LoRA, with a smaller BWT score. For instance, LLaMA-2-7B-Chat's SeqFT OP(BWT) is 48.7 (8.3\%), while LoRASeqFT stands at 12.7 (45.7\%). This suggests that when the focus is primarily on sequential tasks, full parameter fine-tuning should be prioritized over parameter-efficient methods like LoRA.

\begin{table}[t]\vspace{-1cm}
\centering
\setlength\tabcolsep{2pt}
\caption{Comparison of the general language understanding and reasoning abilities. \textcolor{blue}{blue} means increase, while \textcolor{red}{red} means decrease.}

\label{tab:general benchmark}
\resizebox{\textwidth}{!}{
\begin{tabular}{@{}lccccccc@{}}
\toprule
& \multicolumn{1}{c}{\textbf{\begin{tabular}[c]{@{}c@{}}MMLU \\ (factuality)\end{tabular}}} 
& \multicolumn{1}{c}{\textbf{\begin{tabular}[c]{@{}c@{}}GSM\\ (math)\end{tabular}}}  
& \multicolumn{1}{c}{\textbf{\begin{tabular}[c]{@{}c@{}}BBH \\ (reasoning)\end{tabular}}}  
& \multicolumn{1}{c}{\textbf{\begin{tabular}[c]{@{}c@{}}TydiQA\\ (multilinguality)\end{tabular}}} 
& \multicolumn{1}{l}{\textbf{\begin{tabular}[c]{@{}c@{}}BoolQ\\ (comprehension)\end{tabular}}} 
& \multicolumn{1}{l}{\textbf{\begin{tabular}[c]{@{}c@{}}PIQA\\ (commonsense)\end{tabular}} }
& \multicolumn{1}{c}{\textbf{$\Delta R_t^G $}}  \\ 

\cmidrule(l){2-8} 

& \multicolumn{1}{c}{\textbf{\begin{tabular}[c]{@{}c@{}}ACC \\ (5-shot)\end{tabular}}}        
& \multicolumn{1}{c}{\textbf{\begin{tabular}[c]{@{}c@{}}EM\\ (8-shot, CoT)\end{tabular}}} 
& \multicolumn{1}{c}{\textbf{\begin{tabular}[c]{@{}c@{}}EM \\ (3-shot, CoT)\end{tabular}}} 
& \multicolumn{1}{c}{\textbf{\begin{tabular}[c]{@{}c@{}}F1\\ (1-shot, GP)\end{tabular}}}          
& \multicolumn{1}{c}{\textbf{\begin{tabular}[c]{@{}c@{}}ACC\\ (0-shot)\end{tabular}}}       
& \multicolumn{1}{c}{\textbf{\begin{tabular}[c]{@{}c@{}}ACC \\ (0-shot) \end{tabular}}    }    
 & \multicolumn{1}{l}{\textbf{}}         \\ 
 
 \midrule
LLaMA-2-7B-Chat                       &  $46.56$ &  $26.08$ &  $40.23$ &  $23.47$ & $70.55$ &  $76.22$   &  $0$    \\

LLaMA-2-7B-Chat-Seq      & \cellcolor[HTML]{FFF7F7}$46.43$ & \cellcolor[HTML]{FF8F8F}$3.49$ & \cellcolor[HTML]{FFB5B5}$30.11$ & \cellcolor[HTML]{D6D6FF}$33.23$ & \cellcolor[HTML]{DCDCFF}$77.89$ & \cellcolor[HTML]{F7F7FF}$76.5$ & \cellcolor[HTML]{FFDFDF}$-2.58$ \\

LLaMA-2-7B-Chat-LoraSeq & \cellcolor[HTML]{FFD6D6}$42.28$ & \cellcolor[HTML]{FF9F9F}$14.71$ & \cellcolor[HTML]{FFBBBB}$33.61$ & \cellcolor[HTML]{FFF1F1}$21.72$ & \cellcolor[HTML]{FF8A8A}$53.43$ & \cellcolor[HTML]{FFF9F9}$75.19$ & \cellcolor[HTML]{FFB2B2}$-7.03$ \\

LLaMA-2-7B-Chat-Replay & \cellcolor[HTML]{F7F7FF}$47.04$ & \cellcolor[HTML]{FF8888}$3.03$ & \cellcolor[HTML]{FFDADA}$36.61$ & \cellcolor[HTML]{DADAFF}$31.57$ & \cellcolor[HTML]{E0E0FF}$75.75$ & \cellcolor[HTML]{FFF8F8}$75.3$ & \cellcolor[HTML]{FFDCDC}$-2.31$ \\

\midrule
LLaMA-2-13B-Chat                        &  $54.61$ &  $43.14$ &  $49.70$ &  $27.65$ &  $81.5$ &  $78.24$   &  $0$    \\

LLaMA-2-13B-Chat-Seq & \cellcolor[HTML]{FFB0B0}$41.88$ & \cellcolor[HTML]{FF7070}$2.12$ & \cellcolor[HTML]{FF9090}$19.47$ & \cellcolor[HTML]{EAEAFF}$32.27$ & \cellcolor[HTML]{F7F7FF}$82.08$ & \cellcolor[HTML]{FFF8F8}$77.15$ & \cellcolor[HTML]{FFADAD}$-13.32$ \\

LLaMA-2-13B-Chat-LoraSeq & \cellcolor[HTML]{FFDADA}$50.63$ & \cellcolor[HTML]{FF8F8F}$24.72$ & \cellcolor[HTML]{FFB5B5}$38.98$ & \cellcolor[HTML]{FFF8F8}$26.93$ & \cellcolor[HTML]{FFAAAA}$68.96$ & \cellcolor[HTML]{FFFBFB}$78.02$ & \cellcolor[HTML]{FFCACA}$-7.78$ \\

LLaMA-2-13B-Chat-Replay & \cellcolor[HTML]{FFD4D4}$47.72$ & \cellcolor[HTML]{FF7373}$2.96$ & \cellcolor[HTML]{FFAEAE}$36.52$ & \cellcolor[HTML]{EAEAFF}$32.52$ & \cellcolor[HTML]{F7F7FF}$82.45$ & \cellcolor[HTML]{FFF6F6}$76.88$ & \cellcolor[HTML]{FFC6C6}$-9.3$ \\

\midrule
Baichuan2-7B-Instruct                       &  $53.80$ &  $33.21$ &  $35.66$ &  $20.64$ &  $77.09$ &  $74.05$   &  $0$    \\

Baichuan2-7B-Instruct-Seq & \cellcolor[HTML]{FFDDDD}$46.92$ & \cellcolor[HTML]{FF8B8B}$4.25$ & \cellcolor[HTML]{F0F0FF}$37.45$ & \cellcolor[HTML]{B5B5FF}$35.20$ & \cellcolor[HTML]{EDEFFF}$79.08$ & \cellcolor[HTML]{FAFAFF}$74.21$ & \cellcolor[HTML]{FFD7D7}$-2.89$ \\

Baichuan2-7B-Instruct-LoraSeq & \cellcolor[HTML]{FFD9D9}$52.14$ & \cellcolor[HTML]{FFB8B8}$22.74$ & \cellcolor[HTML]{FFCDCD}$27.53$ & \cellcolor[HTML]{D8D8FF}$30.99$ & \cellcolor[HTML]{FFD7D7}$75.23$ & \cellcolor[HTML]{F5F5FF}$74.86$ & \cellcolor[HTML]{FFE5E5}$-1.83$ \\

Baichuan2-7B-Instruct-Replay & \cellcolor[HTML]{FFD6D6}$45.72$ & \cellcolor[HTML]{FFA1A1}$8.19$ & \cellcolor[HTML]{FFFFFF}$35.61$ & \cellcolor[HTML]{B6B6FF}$34.65$ & \cellcolor[HTML]{E8E8FF}$80.06$ & \cellcolor[HTML]{FFEBEB}$72.69$ & \cellcolor[HTML]{FFD7D7}$-2.93$ \\

\midrule
Vicuna-7B-V1.5 &  $51.28$ &  $23.65$ &  $43.32$ & $22.38$ & $78.56$ &  $77.42$   &  $0$    \\

Vicuna-7B-V1.5-Seq & \cellcolor[HTML]{FFEBEB}$49.46$ & \cellcolor[HTML]{FF8E8E}$3.87$ & \cellcolor[HTML]{FFD5D5}$39.25$ & \cellcolor[HTML]{D0D0FF}$33.92$ & \cellcolor[HTML]{FFF7F7}$77.74$ & \cellcolor[HTML]{FFEBEB}$75.73$ & \cellcolor[HTML]{FFDFDF}$-2.77$ \\

Vicuna-7B-V1.5-LoraSeq & \cellcolor[HTML]{FFDCDC}$48.37$ & \cellcolor[HTML]{FFD2D2}$18.89$ & \cellcolor[HTML]{FFA5A5}$28.16$ & \cellcolor[HTML]{F2F2FF}$25.84$ & \cellcolor[HTML]{FFC7C7}$67.24$ & \cellcolor[HTML]{FFF2F2}$76.23$ & \cellcolor[HTML]{FFCFCF}$-5.13$ \\

Vicuna-7B-V1.5-Replay & \cellcolor[HTML]{FFD5D5}$47.20$ & \cellcolor[HTML]{FF9292}$4.78$ & \cellcolor[HTML]{FFD6D6}$39.26$ & \cellcolor[HTML]{D9D9FF}$31.86$ & \cellcolor[HTML]{F9F9FF}$78.92$ & \cellcolor[HTML]{F6F6FF}$80.13$ & \cellcolor[HTML]{FFDEDE}$-2.41$ \\

\midrule
Vicuna-13B-V1.5                       &  $56.16$ &  $36.09$ &  $51.29$ & $24.89$ & $82.45$ &  $78.89$   &  $0$    \\

Vicuna-13B-V1.5-Seq & \cellcolor[HTML]{FFB0B0}$37.93$ & \cellcolor[HTML]{FF8080}$2.81$ & \cellcolor[HTML]{FFBABA}$35.23$ & \cellcolor[HTML]{E0E0FF}$36.86$ & \cellcolor[HTML]{F8F8FF}$83.43$ & \cellcolor[HTML]{FFF7F7}$77.86$ & \cellcolor[HTML]{FFC1C1}$-9.27$ \\

Vicuna-13B-V1.5-LoraSeq & \cellcolor[HTML]{FFDADA}$52.46$ & \cellcolor[HTML]{FF8F8F}$22.14$ & \cellcolor[HTML]{FFB5B5}$41.22$ & \cellcolor[HTML]{F2F2FF}$27.86$ & \cellcolor[HTML]{FF8888}$67.71$ & \cellcolor[HTML]{FFF8F8}$77.53$ & \cellcolor[HTML]{FFD6D6}$-6.18$ \\

Vicuna-13B-V1.5-Replay & \cellcolor[HTML]{FFDADA}$48.73$ & \cellcolor[HTML]{FF8282}$3.11$ & \cellcolor[HTML]{FFD4D4}$42.94$ & \cellcolor[HTML]{D2D2FF}$39.60$ & \cellcolor[HTML]{F5F5FF}$84.71$ & \cellcolor[HTML]{FFF6F6}$77.53$ & \cellcolor[HTML]{FFE5E5}$-5.52$ \\

 \bottomrule
\end{tabular}
}
\end{table}

\begin{figure}[ht]
    \centering
    \begin{subfigure}[b]{0.49\textwidth}
    \label{fig:a}
    \includegraphics[width=\textwidth]{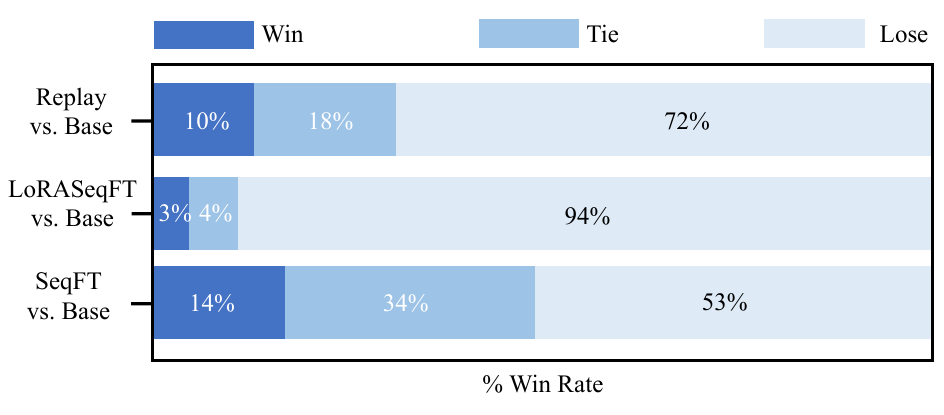}
    \subcaption{Helpful evaluation}
    
    \end{subfigure}
    \begin{subfigure}[b]{0.49\textwidth}
    \label{fig:b}
    \includegraphics[width=\textwidth]{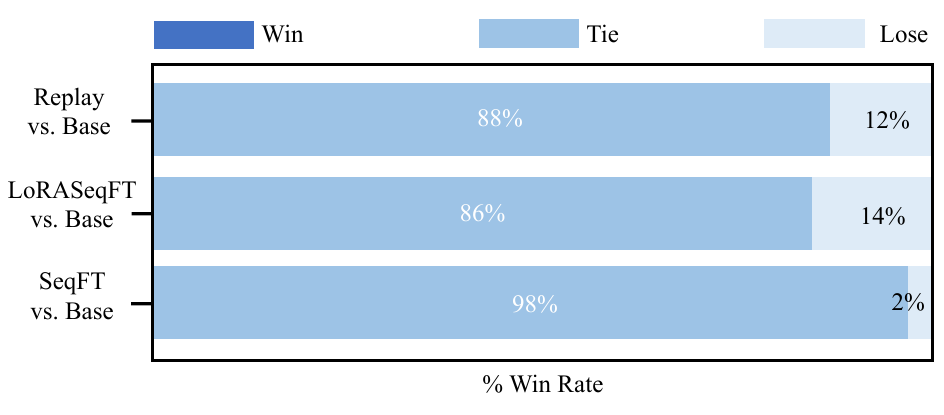}
    \subcaption{Safety evaluation}
        
    \end{subfigure}

    \caption{GPT-4 evaluation with llama-13b-chat, comparing 3 different baselines (Replay, LoRA and Sequential) to the base model across tasks including helpful and safety.}
    \label{fig:3H}
\end{figure}

\subsubsection{Variation of General Ability}
\label{sec:general ability}
Table \ref{tab:general benchmark} presents the evaluations of various LLM models concerning general abilities. The degree of general ability forgetting in LLMs can be analyzed from three perspectives. For a more detailed evaluation, refer to the Appendix.

From the Model Perspective:
\textbf{1)} Nearly all models display a negative General Ability Delta, indicating a general decline in overall capabilities after continual learning.
\textbf{2)} Larger models, in comparison to their smaller counterparts, show a more pronounced forgetting in factual knowledge and reasoning tasks. For instance, the General Ability Delta for llama2-13B-chat-Seq stands at 41.88, whereas the SeqFT version for llama2-7B is 46.43.

From the Task Perspective:
\textbf{1)} Despite the presence of CoT prompts, there is a noticeable decline in math and reasoning abilities across all models, suggesting that these abilities are highly sensitive to new task learning.
\textbf{2)} Excluding the llama2-7b model, most models exhibit a significant drop in performance on MMLU, suggesting a gradual loss of factual knowledge through continual learning.
\textbf{3)} TydiQA task sees a general boost post-training, possibly due to the inclusion of Chinese and German datasets in our sequential tasks. Even more intriguing is the observed enhancement (and some declines) in other languages on TydiQA, suggesting potential cross-linguistic transfer characteristics.
\textbf{4)} Performance shifts on PIQA for most models are subtle, indicating the relative robustness of commonsense knowledge during continual learning.

From the Methodological Perspective:
\textbf{1)} The Replay method proves beneficial in preserving reasoning and factuality skills. Especially for larger models, the mitigation of forgetting through Replay is more pronounced. For instance, for LLaMA-2-7B-Chat, Replay offers a 6.5 EM score boost compared to methods without Replay, while for LLaMA-2-13B-Chat, the increase is 17.1 EM score.

\subsubsection{Instruction Following Ability Analysis}
We evaluated the instruction-following ability of models based on two foundation models: LLaMA-2-7B-Chat and LLaMA-2-13B-Chat. Figure \ref{fig:3H} (a) illustrates the win rate \% for instruction following sequentially trained LLMs and their original versions. Here, the win rate can be approximated as an indicator for the Instruction-following delta. It's evident that all three training methods exhibit a marked decline in instruction-following capabilities compared to their initial versions, with the decline being most pronounced in the LoRA method. Therefore, be cautious when exploring approaches like LoRA for continual learning in LLMs.


\subsubsection{Safety Analysis}
We tested the safety of answers from models LLaMA-2-7B-Chat and LLaMA-2-13B-Chat. Figure \ref{fig:3H} (b) shows the win rate \% for instruction following between the new LLMs and their starting versions. Here, the win rate can be used as a measure for the Safety Delta. Compared to the original models, most answers were rated as 'Tie'. This suggests that the safety of the model's answers is largely unaffected by continual learning on general tasks.

\subsection{Influencing Factors of Forgetting in LLMs}

\subsubsection{Data Quantity \& Training Steps}

\begin{figure}[htbp]
\centering
\begin{minipage}[t]{0.45\textwidth}
\centering
\includegraphics[width=6cm]{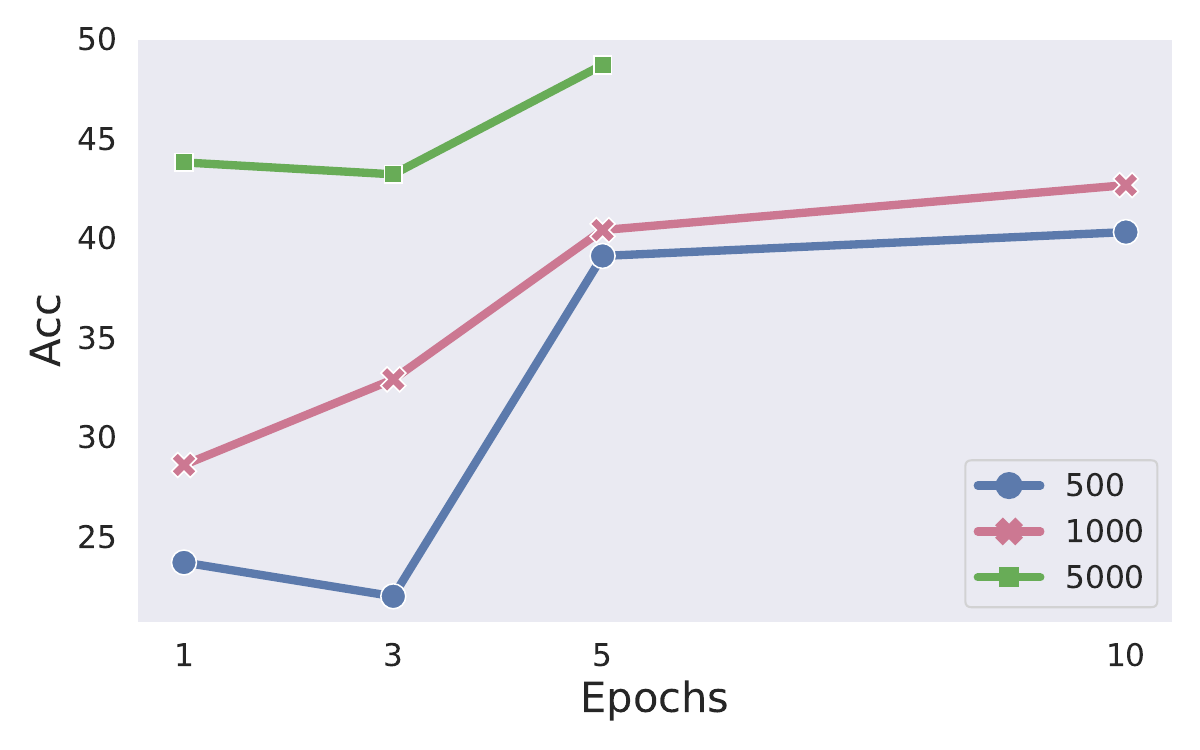}
\caption{Performance evaluation of LLaMA-2-7B-Chat's SeqFT on the TRACE benchmark across varying sample sizes (500, 1000, 5000) and training epochs ({1, 3, 5, 10 (except for 5000)}).}
\label{fig:training_steps}
\end{minipage}
\hspace{.15in}
\begin{minipage}[t]{0.45\textwidth}
\centering
\includegraphics[width=6cm]{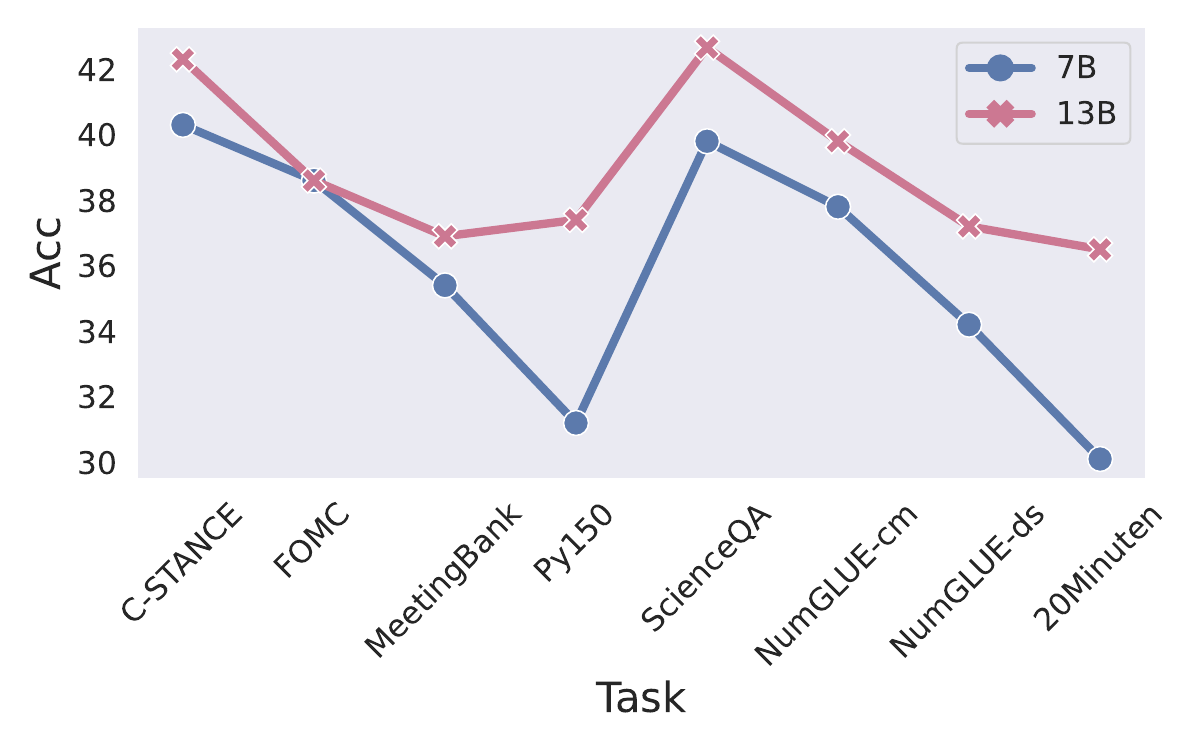}
\caption{Evolution of LLMs' reasoning capabilities post-training on different tasks, measured using the BBH performance metric. We report the results of LLaMA-2-7B-chat and LLaMA-2-13B-chat.}
\label{fig:decoupling}
\end{minipage}
\end{figure}

Figure \ref{fig:training_steps} shows the performance on target tasks for continual learning datasets with different data volumes and training steps. For LLaMA-2-7B-Chat's SeqFT, we tested with 500, 1000, and 5000 samples from each dataset, training them for {1, 3, 5, 10} epochs. Performance improves as data volume grows, indicating at least 5000 samples from the TRACE-selected datasets are needed for full fitting. Additionally, performance improves with up to 5 training epochs, confirming our baseline epoch setting balances target task optimization and retaining existing capabilities.


\subsubsection{Decoupling the Impact of Different Tasks}
From the results in section \ref{sec:general ability}, it's evident that post-training LLMs on our benchmark, their innate reasoning and mathematical abilities see a significant dip. This brings forth the question: How exactly does the reasoning capability of LLMs transform during the continual learning process?

Figure \ref{fig:decoupling} tracks the reasoning ability (assessed via BBH performance) after the completion of each training task. Intriguingly, we observed a surge in the model's reasoning prowess post-training on the ScienceQA task, while it declined for other tasks. Notably, even though the two tasks from NumGLUE are mathematically inclined, their answers don't provide a clear reasoning path. In contrast, ScienceQA does offer such a pathway in its answers. This observation suggests the potential advantage of incorporating reasoning paths during training to preserve and perhaps even enhance the model's reasoning capability.


\section{Reasoning-augmented Continual learning}
Drawing from our earlier findings, which underscored the unique capabilities of LLMs, we were inspired to reconsider how we approach their training. Instead of treating LLMs as traditional models and inundating them with large volumes of data to fit a task's distribution, might we leverage their inherent abilities for rapid task transfer? With these insights as our foundation, we formulated the Reasoning-augmented Continual Learning (RCL) approach. RCL forms reasoning paths on new datasets, aiming to not only preserve LLMs' reasoning capabilities but also to enhance their task transfer and output clarity.

\begin{figure*}[t] \vspace{-1cm}
\centering    \includegraphics[width=0.98\textwidth]{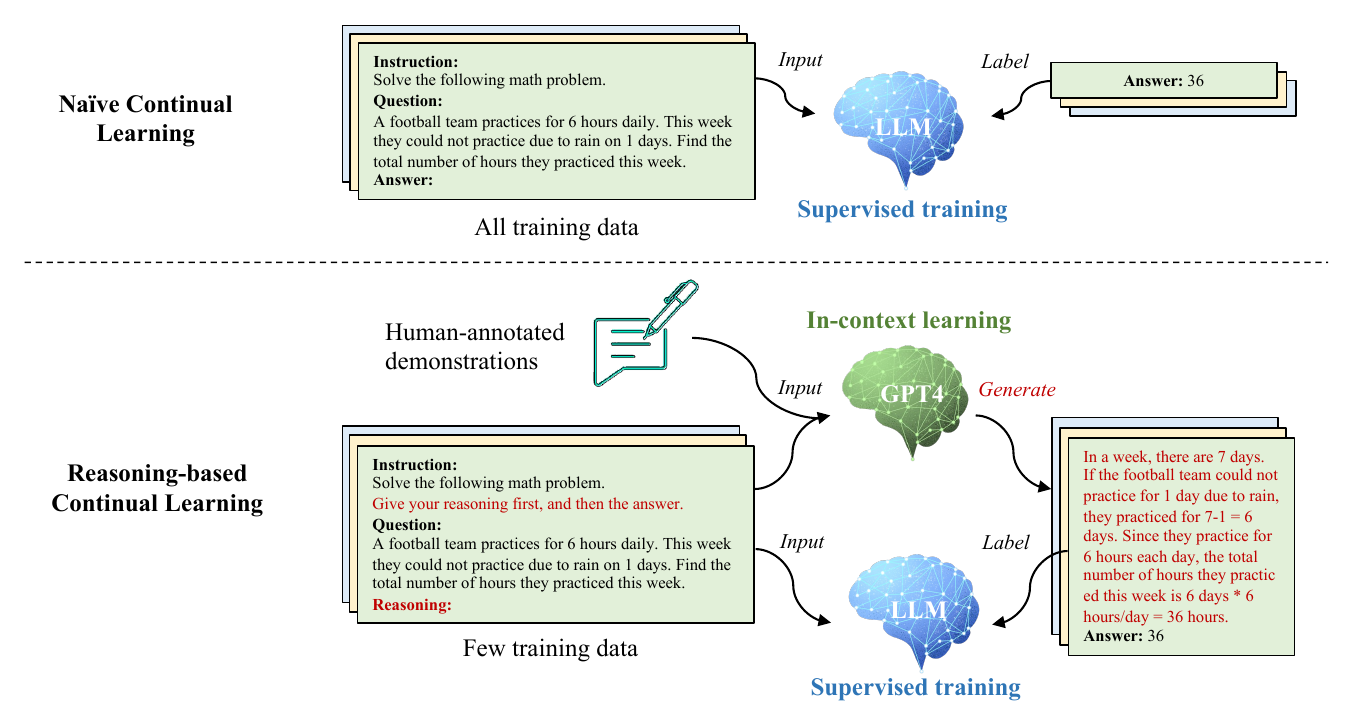} 
    \caption{An overview of Reasoning-augmented continual learning method. Our method unfolds in two stages: 1) Automatic annotation of sample reasoning paths using GPT-4. We guide GPT-4 through in-context learning and validate the generated paths via answer verification. 2) Continual learning on reasoning-augmented dataset.}\label{fig:RCL} 
\end{figure*}

As depicted in Figure \ref{fig:RCL}, RCL has two phases: automated reasoning annotation and sequential training on the augmented dataset. Domain experts created prompts for each task. Three samples per task were manually annotated. GPT-4, using these prompts, generated reasoning paths for every entry. Reasoning was verified against ground truth and underwent human checks. We relied on machine-generated answers due to cost concerns and the consistency of the LM-generated text. To validate reasoning quality, we manually inspected outputs, achieving a 94\% approval rate on a 100-sample check, highlighting GPT-4's reliability. Following this, supervised training was conducted on the target LLM, keeping hyperparameter settings consistent with baselines.

\subsection{Performance on Target Sequential Tasks}
Table \ref{RCL:Baseline} provides a side-by-side comparison of the performance of our RCL method against other techniques, using LLaMA-2-7B-Chat as the foundational model and limiting training samples to 500 for each task. Through an ablation study contrasting single-task training (SingleFT) with multi-task training, and assessing the impact of reasoning-augmented data, we observed that integrating reasoning pathways into the data consistently boosts performance over the original dataset.

Moreover, our approach, even when trained with just 500 samples, achieves comparable results to the SeqFT method trained on 5000 samples. Furthermore, by leveraging fewer datasets and training steps in our method, we would mitigate the decline in LLMs' inherent capabilities.

\begin{wraptable}{r}{0pt}
\hfill
\begin{minipage}[H]{0.45\textwidth}
\centering
\vspace{-1cm}
\captionof{table}{OP and BWT for different baselines. \textbf{O-Lora}\citep{wang-etal-2023-o-lora} and \textbf{PP}\citep{razdaibiedina2023progressive} refer to two SoTA continual learning methods. All methods except \textbf{Seq(5k)} and \textbf{ICL} are training with 0.5k samples. \textbf{Re} refers to reasoning-augmented, \textbf{Single FT} refers to fine-tuning the model on single task and \textbf{MT} refers to Multi-task training.}
\label{RCL:Baseline}
\scalebox{1}{
\begin{tabular}{c|c|c}
    \hline
          &Overall avg. & BWT  \\
          \hline
         ICL& $39.5$& -\\ 
         ICL+Re &$41.1$&- \\
         O-Lora &$41.3$& $6.2\%$\\
         PP & $46.2$ & 2.3\%\\
         SeqFT &$23.0$& $19\%$\\
         SeqFT(5k) &$\textbf{48.7}$& $8.3\%$\\
         \cellcolor[HTML]{90EE90}RCL &$46.6$& $13\%$\\
         \hline
         SingleFT &$57.6$& - \\
         SingleFT+Re &$58.1$& - \\
         MT w/o. Re &$52.3$&- \\
         MT w. Re &$58.2$&- \\
    \hline
    \end{tabular}
}\vspace{-1cm}
\end{minipage}
\end{wraptable}

\subsection{Impacts on General Ability}
\begin{figure}[t] \vspace{-1.2cm}
    \centering
    \includegraphics[width=\textwidth]{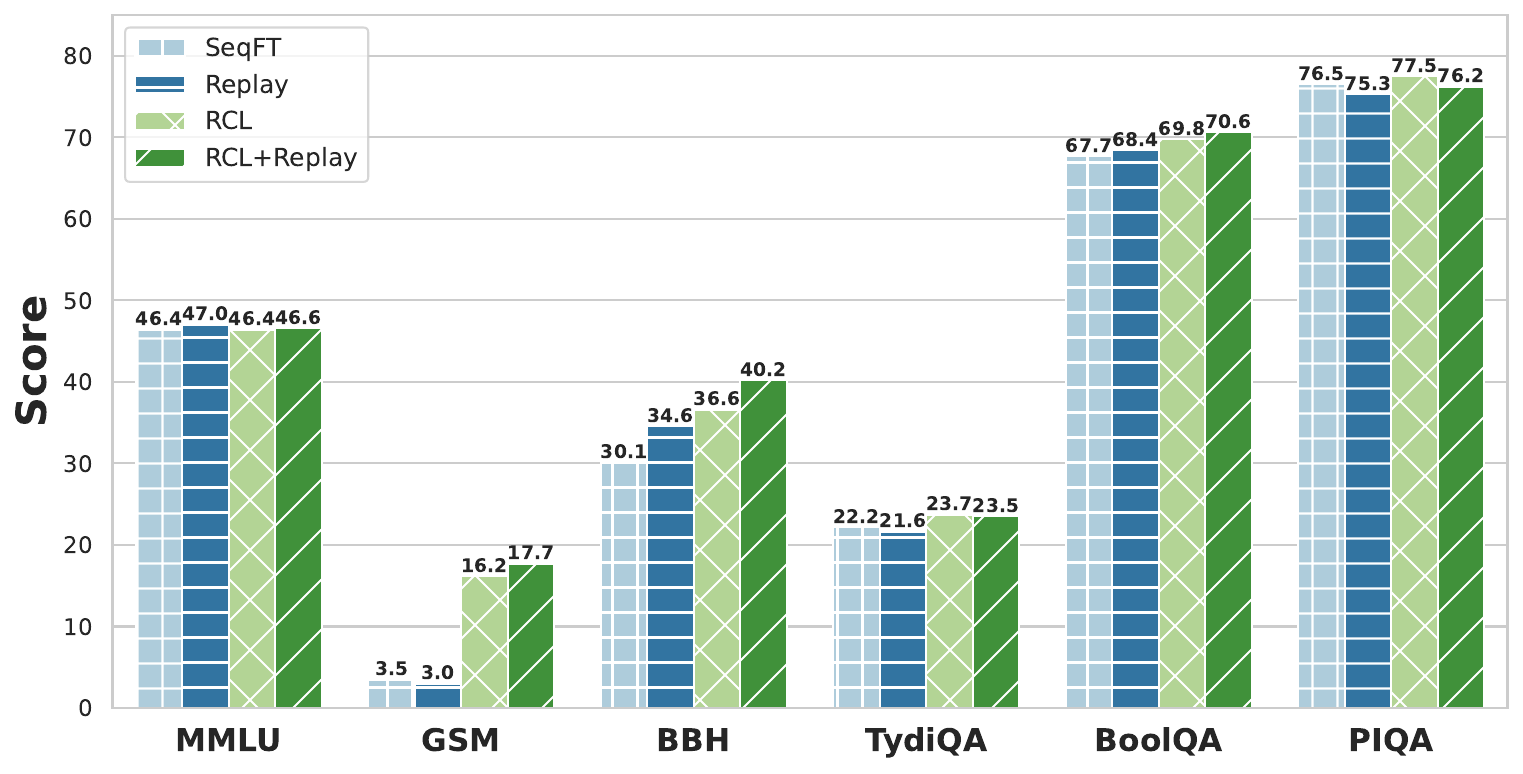}
    \caption{OpenCompass Evaluation Results.\textbf{RCL+Replay} refers to combining our RCL method with replay method.}
    \label{fig:RCL_reasoning}
\end{figure}

We report the performance of general abilities in Figure \ref{fig:RCL_reasoning}. We can conclude that RCL reaches comparable performance with SeqFT and Replay method on MMLU, TydiQA, BoolQA and PIQA though only. However, RCL stands out in reasoning tasks such as GSM and BBH. For instance, for the GSM task, RCL outperforms SeqFT and Replay by 12.7 and 13.2 points respectively, showing the advantages of providing reasoning paths in maintaining the reasoning abilities of models. Besides, combining RCL with replay further improves its performance on reasoning tasks.

\subsection{Impacts on Instruction-Following and Safety}
The impact of incorporating RCL on instruction-following capabilities is presented in Table \ref{tab:RCL_instruction_follow}. It's evident that RCL enhances the model's ability to follow instructions by 8\% and 5\% compared to SeqFT and Replay, respectively.

\section{Discussion}

\textbf{Can traditional continual learning methods be effectively applied to LLMs?}

In section \ref{Related Work}, we introduce various traditional continual learning methods, including replay-based, regularization-based, and architecture-based approaches.
Unfortunately, several characteristics of LLMs challenge the straightforward adoption of these approaches:
\begin{enumerate}[leftmargin=*]
\item \textbf{High Training Cost}: LLMs require significant data for both pre-training and alignment, leading to a high training cost. Using simple replay to maintain past capabilities can be very expensive. Therefore, selecting key data from past training to keep LLMs' diverse predictive abilities is essential.

\item \textbf{Large Number of Parameters}: The huge parameter size of LLMs demands advanced hardware for training. Many regularization techniques \citep{kirkpatrick2017overcoming,farajtabar2020orthogonal,lopez2017gradient}
 need to store gradients from past tasks, which is a big challenge for both CPU and GPU memory.

\item \textbf{One-for-All Deployment of LLMs}: LLMs are designed for a wide range of tasks, meaning tailoring parameters for specific tasks might limit their ability to generalize to new tasks. Additionally, methods that adjust the network dynamically can complicate deployment, as it becomes tricky to handle multiple task queries at once.
\end{enumerate}

\textbf{How should LLMs approach continual learning?}

Our experiments with TRACE show that direct end-to-end training might compel LLMs to focus myopically on specific patterns of the target task, such as shortcuts, thereby undermining their capacities in more universal scenarios.
Intrinsically, LLMs are trained on large and varied datasets. So, they already have the skills to handle many tasks, and can even learn with very few examples. 
Based on the ideas from LIMA's Superficial Alignment Hypothesis \citep{zhou2023lima}, our focus should perhaps pivot more towards adeptly aligning LLMs' existing capabilities to novel tasks rather than embarking on learning from scratch. 
Consequently, strategies like our RCL approach, which capitalize on the LLMs' inherent abilities for quick transfer to new tasks, might also serve as potent tools in mitigating catastrophic forgetting.

\section{Conclusion}
Existing benchmarks often fall short in thoroughly evaluating LLMs, either due to their simplistic nature or neglect of critical capabilities like instruction following and safety.
To tackle this, we introduced TRACE, a comprehensive benchmark with diverse challenging tasks and well-rounded metrics. Our experiments showed the real challenges LLMs face, especially a clear drop in their general abilities during continual learning. At the same time, our Reasoning-augmented Continual Learning (RCL) method highlights the importance of using reasoning in training, even though it's not a complete solution. We believe this area is very important and hope our work lays a solid foundation for future studies.



\bibliography{iclr2024_conference}
\bibliographystyle{iclr2024_conference}

\appendix
\section*{Appendix}

\subsection{Implementation Details}
\label{sec:imple details}
During the training phase, for the baselines without LoRA adapters, we consistently trained with 5000 samples with a constant learning rate of 1e-5. For the datasets we used (C-STANCE, FOMC, MeetingBank, Py150, ScienceQA, NumGLUE-cm, NumGLUE-ds, 20Minuten), we train for 1, 1, 5, 5, 1, 5, 5, 5 epochs respectively. While for the baselines with LoRA adapters, we trained with 5000 samples with a constant learning rate of 1e-4 for 5, 3, 7, 5, 3, 5, 5, 7 epochs respectively. Our training settings incorporated a weight decay set to 0, and a batch size of 128.
For the testing phase, we use a temperature of 0.1.
All our training and inference experiments were conducted on a machine equipped with 8x80G Nvidia A100, and were implemented using DeepSpeed repository.
All models are trained on 8 A100 GPUs with 80G memory with full parameters fine-tuning. We leave the exploration of large models like LLaMa-2-65B-Chat for future work due to the current hardware limits.

All general benchmark evaluations were conducted using the Open-Compass toolkit \citep{2023opencompass}, adopting its default configuration.


\subsection{Trace dataset statistics}
In this section, we represent the overview of dataset statistics, including source, average lenght, metric, language and number of samples of each dataset in TRACE benchmark.
\begin{table}[H]
\caption{An overview of dataset statistics in TRACE. 
'Source' indicates the context's origin. 'Avg len' represents word count for English, German, and code datasets, and character count for Chinese. 'SARI' is a score specific to simplification.}
\begin{tabular}{@{}llllll@{}}
\toprule
Dataset                         & Source       & Avg len & Metric          & Language & \#data \\ \midrule
\textit{Domain-specific}        &              &         &                 &          &        \\
ScienceQA                       & Science      & 210     & Accuracy        & English  & 5,000  \\
FOMC                            & Finance      & 51      & Accuracy        & English  & 5,000  \\
MeetingBank                     & Meeting      & 2853    & ROUGE-L         & English  & 5,000  \\ \midrule
\textit{Multi-lingual}          &              &         &                 &          &        \\
C-STANCE                        & Social media & 127     & Accuracy        & Chinese  & 5,000  \\
20Minuten                       & News         & 382     & SARI            & Germany  & 5,000  \\ \midrule
\textit{Code completion}        &              &         &                 &          &        \\
Py150                           & Github       & 422     & Edim similarity & Python   & 5,000  \\ \midrule
\textit{Mathematical reasoning} &              &         &                 &          &        \\
NumGLUE-cm                      & Math         & 32      & Accuracy        & English  & 5,000  \\
NumGLUE-ds                      & Math         & 21      & Accuracy        & English  & 5,000  \\ \bottomrule
\end{tabular}
\label{Dataset statis}
\end{table}

\subsection{Supplementary Tables}

\begin{table}[H]
    \centering
    \caption{Instruction-following abilities of SeqFT, Replay and RCL}
    \begin{tabular}{c|c|c|c}
        \hline
         &Win&Tie&Loss  \\
         \hline
        SeqFT & 12\% & 30\% & 58\%  \\ 
        Replay & 15\% & 31\% & 55\% \\
        RCL & 20\% & 30\% & 50\% \\ 
        \hline
    \end{tabular}

    \label{tab:RCL_instruction_follow}
\end{table}

\subsection{Detailed Experiments Results}
In this section, we report the detailed experiment results in our paper. The model includes Baichuan-7b,LLaMA2-7b-chat, LLaMA2-13b-chat, Vicuna-7b and Vicuna-13b. The results are shown in Table \ref{Table:6}~\ref{Tabel:27}.

\subsubsection{In-Context Learning}
Table \ref{Table:6} represents the performance of different models with in-context learning.  
\begin{table}[H]
\centering
\caption{Detailed results of in-context learning of different large language models.}
\begin{tabular}{@{}llllll@{}}
\toprule
Task/Model  & Baichuan-7b & LLaMA2-7b-chat               & LLaMA2-13b-chat & Vicuna-7b & Vicuna-13b                  \\ \midrule
C-STANCE    & 0.58        & 0.4                          & 0.366           & 0.403     & 0.57                        \\
FOMC        & 0.63        & 0.483                        & 0.519           & 0.551     & {\color[HTML]{1F2329} 0.61} \\
MeetingBank & 0.225       & 0.198                        & 0.221           & 0.223     & 0.229                       \\
Py150       & 0.586       & 0.522                        & 0.539           & 0.529     & 0.585                       \\
ScienceQA   & 0.68        & {\color[HTML]{1F2329} 0.628} & 0.689           & 0.695     & 0.7                         \\
NumGLUE-cm  & 0.271       & 0.284                        & 0.407           & 0.284     & 0.347                       \\
NumGLUE-ds  & 0.23        & 0.203                        & 0.218           & 0.302     & 0.33                        \\
20Minuten   & 0.366       & 0.395                        & 0.395           & 0.392     & 0.378                       \\ \midrule
average     & 0.446       & 0.389                        & 0.419           & 0.422     & 0.469                       \\ \bottomrule
\end{tabular}
\label{Table:6}
\end{table}

\subsubsection{SeqFT method}
Table \ref{Table:7} - \ref{Table:11} shows the detailed performance of different models of each round during the continual learning. SeqFT represents sequential fine-tuning.
\begin{table}[H]
\caption{Detailed results of continual learning of Baichuan-7b.}
\centering
\begin{tabular}{@{}lllllllll@{}}
\toprule
Task\textbackslash{}Round & 1    & 2     & 3     & 4     & 5                           & 6     & 7     & 8     \\ \midrule
C-STANCE                  & 0.62 & 0.629 & 0.663 & 0.621 & 0.531                       & 0.55  & 0.588 & 0.579 \\
FOMC                      & -    & 0.681 & 0.353 & 0.318 & {\color[HTML]{1F2329} 0.02} & 0.363 & 0.347 & 0.335 \\
MeetingBank               & -    & -     & 0.442 & 0.351 & 0.371                       & 0.379 & 0.389 & 0.364 \\
Py150                     & -    & -     & -     & 0.626 & 0.562                       & 0.586 & 0.589 & 0.58  \\
ScienceQA                 & -    & -     & -     & -     & 0.77                        & 0.68  & 0.5   & 0.44  \\
NumGLUE-cm                & -    & -     & -     & -     & -                           & 0.358 & 0.247 & 0.284 \\
NumGLUE-ds                & -    & -     & -     & -     & -                           & -     & 0.64  & 0.475 \\
20Minuten                 & -    & -     & -     & -     & -                           & -     & -     & 0.415 \\ \midrule
average                   &      &       &       &       &                             &       &       & 0.434 \\
BWT                       &      &       &       &       &                             &       &       & -0.154   \\ \bottomrule
\end{tabular}

\label{Table:7}
\end{table}

\begin{table}[H]
\caption{Detailed results of continual learning of LLaMA-7b-chat.}
\centering
\begin{tabular}{@{}lllllllll@{}}
\toprule
Task\textbackslash{}Round & 1   & 2     & 3     & 4     & 5                        & 6     & 7     & 8     \\ \midrule
C-STANCE                  & 0.5 & 0.456 & 0.448 & 0.453 & 0.435                    & 0.442 & 0.436 & 0.454 \\
FOMC                      & -   & 0.735 & 0.67  & 0.658 & {\color[HTML]{1F2329} 0} & 0.595 & 0.577 & 0.609 \\
MeetingBank               & -   & -     & 0.523 & 0.459 & 0.433                    & 0.446 & 0.442 & 0.457 \\
Py150                     & -   & -     & -     & 0.58  & 0.459                    & 0.509 & 0.508 & 0.512 \\
ScienceQA                 & -   & -     & -     & -     & 0.764                    & 0.636 & 0.45  & 0.637 \\
NumGLUE-cm                & -   & -     & -     & -     & -                        & 0.383 & 0.247 & 0.272 \\
NumGLUE-ds                & -   & -     & -     & -     & -                        & -     & 0.582 & 0.548 \\
20Minuten                 & -   & -     & -     & -     & -                        & -     & -     & 0.408 \\ \midrule
average                   &     &       &       &       &                          &       &       & 0.487 \\
BWT                       &     &       &       &       &                          &       &       & -0.083   \\ \bottomrule
\end{tabular}
\end{table}

\begin{table}[H]
\caption{Detailed results of continual learning of LLaMA-13b-chat.}

\centering
\begin{tabular}{@{}lllllllll@{}}
\toprule
Task\textbackslash{}Round & 1     & 2     & 3     & 4     & 5                           & 6     & 7     & 8     \\ \midrule
C-STANCE                  & 0.469 & 0.465 & 0.47  & 0.477 & 0.463                       & 0.466 & 0.45  & 0.5   \\
FOMC                      & -     & 0.754 & 0.738 & 0.748 & {\color[HTML]{1F2329} 0.03} & 0.721 & 0.71  & 0.717 \\
MeetingBank               & -     & -     & 0.533 & 0.51  & 0.375                       & 0.421 & 0.385 & 0.351 \\
Py150                     & -     & -     & -     & 0.568 & 0.538                       & 0.537 & 0.541 & 0.547 \\
ScienceQA                 & -     & -     & -     & -     & 0.8                         & 0.655 & 0.241 & 0.55  \\
NumGLUE-cm                & -     & -     & -     & -     & -                           & 0.333 & 0.284 & 0.296 \\
NumGLUE-ds                & -     & -     & -     & -     & -                           & -     & 0.618 & 0.622 \\
20Minuten                 & -     & -     & -     & -     & -                           & -     & -     & 0.408 \\ \midrule
average                   &       &       &       &       &                             &       &       & 0.499 \\
BWT                       &       &       &       &       &                             &       &       & -0.07   \\ \bottomrule
\end{tabular}
\end{table}

\begin{table}[H]
\caption{Detailed results of continual learning of Vicuna-7b.}
\centering
\begin{tabular}{@{}lllllllll@{}}
\toprule
Task\textbackslash{}Round & 1     & 2     & 3     & 4     & 5                        & 6     & 7     & 8     \\ \midrule
C-STANCE                  & 0.532 & 0.451 & 0.439 & 0.448 & 0.149                    & 0.477 & 0.47  & 0.476 \\
FOMC                      & -     & 0.738 & 0.732 & 0.744 & {\color[HTML]{1F2329} 0} & 0.605 & 0.567 & 0.675 \\
MeetingBank               & -     & -     & 0.519 & 0.447 & 0.443                    & 0.427 & 0.417 & 0.439 \\
Py150                     & -     & -     & -     & 0.577 & 0.384                    & 0.486 & 0.482 & 0.482 \\
ScienceQA                 & -     & -     & -     & -     & 0.773                    & 0.7   & 0.608 & 0.649 \\
NumGLUE-cm                & -     & -     & -     & -     & -                        & 0.407 & 0.247 & 0.296 \\
NumGLUE-ds                & -     & -     & -     & -     & -                        & -     & 0.578 & 0.517 \\
20Minuten                 & -     & -     & -     & -     & -                        & -     & -     & 0.403 \\ \midrule
average                   &       &       &       &       &                          &       &       & 0.492 \\
BWT                       &       &       &       &       &                          &       &       & -0.084   \\ \bottomrule
\end{tabular}
\end{table}

\begin{table}[H]
\caption{Detailed results of continual learning of Vicuna-13b.}
\centering
\begin{tabular}{@{}lllllllll@{}}
\toprule
Task\textbackslash{}Round & 1     & 2     & 3     & 4     & 5                        & 6     & 7     & 8     \\ \midrule
C-STANCE                  & 0.527 & 0.43  & 0.471 & 0.497 & 0.374                    & 0.468 & 0.469 & 0.484 \\
FOMC                      & -     & 0.741 & 0.739 & 0.731 & {\color[HTML]{1F2329} 0} & 0.754 & 0.678 & 0.714 \\
MeetingBank               & -     & -     & 0.549 & 0.532 & 0.53                     & 0.491 & 0.427 & 0.412 \\
Py150                     & -     & -     & -     & 0.564 & 0.54                     & 0.546 & 0.538 & 0.552 \\
ScienceQA                 & -     & -     & -     & -     & 0.79                     & 0.616 & 0.586 & 0.633 \\
NumGLUE-cm                & -     & -     & -     & -     & -                        & 0.346 & 0.309 & 0.358 \\
NumGLUE-ds                & -     & -     & -     & -     & -                        & -     & 0.622 & 0.572 \\
20Minuten                 & -     & -     & -     & -     & -                        & -     & -     & 0.41  \\ \midrule
average                   &       &       &       &       &                          &       &       & 0.517 \\
BWT                       &       &       &       &       &                          &       &       & -0.059   \\ \bottomrule
\end{tabular}
\label{Table:11}
\end{table}

\subsubsection{SeqLoraFT method}
Table 12~16 shows the detailed performance of different models of each round during the continual learning. SeqLoRAFT represents sequential fine-tuning with LoRA adapters.
\begin{table}[H]
\caption{Detailed results of continual learning of Baichuan-7b with LoRA adapters.}
\centering
\begin{tabular}{@{}lllllllll@{}}
\toprule
Task\textbackslash{}Round & 1     & 2     & 3     & 4     & 5                            & 6     & 7     & 8     \\ \midrule
C-STANCE                  & 0.613 & 0.601 & 0.597 & 0.584 & 0.506                        & 0.504 & 0.53  & 0.477 \\
FOMC                      & -     & 0.652 & 0.604 & 0.591 & {\color[HTML]{1F2329} 0.602} & 0.588 & 0.587 & 0.417 \\
MeetingBank               & -     & -     & 0.345 & 0.334 & 0.333                        & 0.343 & 0.34  & 0.337 \\
Py150                     & -     & -     & -     & 0.588 & 0.472                        & 0.539 & 0.517 & 0.472 \\
ScienceQA                 & -     & -     & -     & -     & 0.641                        & 0.68  & 0.625 & 0.63  \\
NumGLUE-cm                & -     & -     & -     & -     & -                            & 0.457 & 0.432 & 0.407 \\
NumGLUE-ds                & -     & -     & -     & -     & -                            & -     & 0.43  & 0.36  \\
20Minuten                 & -     & -     & -     & -     & -                            & -     & -     & 0.407 \\ \midrule
average                   &       &       &       &       &                              &       &       & 0.438 \\
BWT                       &       &       &       &       &                              &       &       & -0.090   \\ \bottomrule
\end{tabular}
\end{table}

\begin{table}[H]
\caption{Detailed results of continual learning of LLaMA-7b-chat with LoRA adapters.}
\centering
\begin{tabular}{@{}lllllllll@{}}
\toprule
Task\textbackslash{}Round & 1     & 2     & 3     & 4     & 5                        & 6     & 7     & 8     \\ \midrule
C-STANCE                  & 0.511 & 0.45  & 0.412 & 0.373 & 0.133                    & 0.391 & 0.294 & 0.277 \\
FOMC                      & -     & 0.713 & 0.55  & 0.452 & {\color[HTML]{1F2329} 0} & 0.421 & 0.341 & 0.24  \\
MeetingBank               & -     & -     & 0.51  & 0.212 & 0.151                    & 0.067 & 0.037 & 0.121 \\
Py150                     & -     & -     & -     & 0.578 & 0.004                    & 0.495 & 0.452 & 0.004 \\
ScienceQA                 & -     & -     & -     & -     & 0.68                     & 0.645 & 0.535 & 0     \\
NumGLUE-cm                & -     & -     & -     & -     & -                        & 0.37  & 0.235 & 0     \\
NumGLUE-ds                & -     & -     & -     & -     & -                        & -     & 0.486 & 0     \\
20Minuten                 & -     & -     & -     & -     & -                        & -     & -     & 0.37  \\ \midrule
average                   &       &       &       &       &                          &       &       & 0.127 \\
BWT                       &       &       &       &       &                          &       &       & -0.457   \\ \bottomrule
\end{tabular}
\end{table}

\begin{table}[H]
\caption{Detailed results of continual learning of LLaMA-13b with LoRA adapters.}
\centering
\begin{tabular}{@{}lllllllll@{}}
\toprule
Task\textbackslash{}Round & 1    & 2     & 3     & 4     & 5                           & 6     & 7     & 8     \\ \midrule
C-STANCE                  & 0.62 & 0.36  & 0.432 & 0.491 & 0.18                        & 0.42  & 0.411 & 0.124 \\
FOMC                      & -    & 0.743 & 0.681 & 0.63  & {\color[HTML]{1F2329} 0.53} & 0.605 & 0.579 & 0     \\
MeetingBank               & -    & -     & 0.484 & 0.264 & 0.201                       & 0.147 & 0.032 & 0.122 \\
Py150                     & -    & -     & -     & 0.581 & 0.397                       & 0.488 & 0.497 & 0.249 \\
ScienceQA                 & -    & -     & -     & -     & 0.75                        & 0.729 & 0.714 & 0.68  \\
NumGLUE-cm                & -    & -     & -     & -     & -                           & 0.58  & 0.296 & 0.259 \\
NumGLUE-ds                & -    & -     & -     & -     & -                           & -     & 0.62  & 0.386 \\
20Minuten                 & -    & -     & -     & -     & -                           & -     & -     & 0.417 \\ \midrule
average                   &      &       &       &       &                             &       &       & 0.28  \\
BWT                       &      &       &       &       &                             &       &       & -0.365   \\ \bottomrule
\end{tabular}
\end{table}

\begin{table}[H]
\caption{Detailed results of continual learning of Vicuna-7b with LoRA adapters.}
\centering
\begin{tabular}{@{}lllllllll@{}}
\toprule
Task\textbackslash{}Round & 1     & 2     & 3     & 4     & 5                        & 6     & 7     & 8     \\ \midrule
C-STANCE                  & 0.514 & 0.452 & 0.433 & 0.446 & 0                        & 0.344 & 0.089 & 0.141 \\
FOMC                      & -     & 0.715 & 0.48  & 0.427 & {\color[HTML]{1F2329} 0} & 0.272 & 0.304 & 0.29  \\
MeetingBank               & -     & -     & 0.5   & 0.113 & 0.144                    & 0.026 & 0.011 & 0.07  \\
Py150                     & -     & -     & -     & 0.573 & 0.222                    & 0.47  & 0.452 & 0.413 \\
ScienceQA                 & -     & -     & -     & -     & 0.67                     & 0.632 & 0.53  & 0.6   \\
NumGLUE-cm                & -     & -     & -     & -     & -                        & 0.407 & 0.37  & 0.259 \\
NumGLUE-ds                & -     & -     & -     & -     & -                        & -     & 0.545 & 0.492 \\
20Minuten                 & -     & -     & -     & -     & -                        & -     & -     & 0.409 \\ \midrule
average                   &       &       &       &       &                          &       &       & 0.334 \\
BWT                       &       &       &       &       &                          &       &       & -0.237   \\ \bottomrule
\end{tabular}
\end{table}

\begin{table}[H]
\caption{Detailed results of continual learning of Vicuna-13b with LoRA adapters.}
\centering
\begin{tabular}{@{}lllllllll@{}}
\toprule
Task\textbackslash{}Round & 1     & 2     & 3     & 4     & 5                            & 6     & 7     & 8     \\ \midrule
C-STANCE                  & 0.524 & 0.504 & 0.394 & 0.385 & 0.389                        & 0.347 & 0.329 & 0.07  \\
FOMC                      & -     & 0.74  & 0.68  & 0.616 & {\color[HTML]{1F2329} 0.188} & 0.62  & 0.438 & 0.04  \\
MeetingBank               & -     & -     & 0.495 & 0.24  & 0.157                        & 0.132 & 0.08  & 0.14  \\
Py150                     & -     & -     & -     & 0.6   & 0.368                        & 0.52  & 0.491 & 0.256 \\
ScienceQA                 & -     & -     & -     & -     & 0.77                         & 0.75  & 0.732 & 0.74  \\
NumGLUE-cm                & -     & -     & -     & -     & -                            & 0.407 & 0.346 & 0.346 \\
NumGLUE-ds                & -     & -     & -     & -     & -                            & -     & 0.569 & 0.52  \\
20Minuten                 & -     & -     & -     & -     & -                            & -     & -     & 0.413 \\ \midrule
average                   &       &       &       &       &                              &       &       & 0.316 \\
BWT                       &       &       &       &       &                              &       &       & -0.284   \\ \bottomrule
\end{tabular}
\end{table}

\subsubsection{Replay method}
Table 17~21 shows the detailed performance of different models of each round during the continual learning with replay data.
\begin{table}[H]
\caption{Detailed results of continual learning of Baichuan-7b with replay data.}
\centering
\begin{tabular}{@{}lllllllll@{}}
\toprule
Task\textbackslash{}Round & 1    & 2    & 3     & 4     & 5                           & 6     & 7     & 8     \\ \midrule
C-STANCE                  & 0.57 & 0.55 & 0.56  & 0.63  & 0.6                         & 0.64  & 0.62  & 0.61  \\
FOMC                      & -    & 0.69 & 0.64  & 0.64  & {\color[HTML]{1F2329} 0.65} & 0.65  & 0.66  & 0.61  \\
MeetingBank               & -    & -    & 0.445 & 0.457 & 0.449                       & 0.466 & 0.461 & 0.482 \\
Py150                     & -    & -    & -     & 0.546 & 0.577                       & 0.577 & 0.613 & 0.583 \\
ScienceQA                 & -    & -    & -     & -     & 0.58                        & 0.51  & 0.54  & 0.57  \\
NumGLUE-cm                & -    & -    & -     & -     & -                           & 0.321 & 0.346 & 0.333 \\
NumGLUE-ds                & -    & -    & -     & -     & -                           & -     & 0.5   & 0.55  \\
20Minuten                 & -    & -    & -     & -     & -                           & -     & -     & 0.405 \\ \midrule
average                   &      &      &       &       &                             &       &       & 0.517 \\
BWT                       &      &      &       &       &                             &       &       & 0.011   \\ \bottomrule
\end{tabular}
\end{table}

\begin{table}[H]
\caption{Detailed results of continual learning of LLaMA-7b-chat with replay data.}
\centering
\begin{tabular}{@{}lllllllll@{}}
\toprule
Task\textbackslash{}Round & 1     & 2     & 3     & 4     & 5                            & 6     & 7     & 8     \\ \midrule
C-STANCE                  & 0.471 & 0.487 & 0.485 & 0.5   & 0.486                        & 0.475 & 0.493 & 0.5   \\
FOMC                      & -     & 0.734 & 0.769 & 0.785 & {\color[HTML]{1F2329} 0.807} & 0.781 & 0.785 & 0.8   \\
MeetingBank               & -     & -     & 0.499 & 0.496 & 0.507                        & 0.494 & 0.492 & 0.51  \\
Py150                     & -     & -     & -     & 0.543 & 0.561                        & 0.546 & 0.552 & 0.55  \\
ScienceQA                 & -     & -     & -     & -     & 0.763                        & 0.78  & 0.78  & 0.785 \\
NumGLUE-cm                & -     & -     & -     & -     & -                            & 0.358 & 0.309 & 0.37  \\
NumGLUE-ds                & -     & -     & -     & -     & -                            & -     & 0.486 & 0.52  \\
20Minuten                 & -     & -     & -     & -     & -                            & -     & -     & 0.406 \\ \midrule
average                   &       &       &       &       &                              &       &       & 0.555 \\
BWT                       &       &       &       &       &                              &       &       & 0.026   \\ \bottomrule
\end{tabular}
\end{table}

\begin{table}[H]
\caption{Detailed results of continual learning of LLaMA-13b-chat with replay data.}
\centering
\begin{tabular}{@{}lllllllll@{}}
\toprule
Task\textbackslash{}Round & 1   & 2     & 3     & 4     & 5                            & 6     & 7     & 8     \\ \midrule
C-STANCE                  & 0.5 & 0.496 & 0.497 & 0.493 & 0.52                         & 0.503 & 0.5   & 0.51  \\
FOMC                      & -   & 0.778 & 0.803 & 0.805 & {\color[HTML]{1F2329} 0.792} & 0.789 & 0.785 & 0.813 \\
MeetingBank               & -   & -     & 0.484 & 0.495 & 0.515                        & 0.499 & 0.503 & 0.482 \\
Py150                     & -   & -     & -     & 0.523 & 0.549                        & 0.532 & 0.534 & 0.523 \\
ScienceQA                 & -   & -     & -     & -     & 0.816                        & 0.8   & 0.804 & 0.792 \\
NumGLUE-cm                & -   & -     & -     & -     & -                            & 0.358 & 0.407 & 0.396 \\
NumGLUE-ds                & -   & -     & -     & -     & -                            & -     & 0.628 & 0.606 \\
20Minuten                 & -   & -     & -     & -     & -                            & -     & -     & 0.407 \\ \midrule
average                   &     &       &       &       &                              &       &       & 0.566 \\
BWT                       &     &       &       &       &                              &       &       & 0.004   \\ \bottomrule
\end{tabular}
\end{table}

\begin{table}[H]
\caption{Detailed results of continual learning of Vicuna-7b with replay data.}
\centering
\begin{tabular}{@{}lllllllll@{}}
\toprule
Task\textbackslash{}Round & 1   & 2     & 3     & 4     & 5                            & 6     & 7     & 8     \\ \midrule
C-STANCE                  & 0.5 & 0.528 & 0.512 & 0.519 & 0.518                        & 0.519 & 0.515 & 0.524 \\
FOMC                      & -   & 0.747 & 0.803 & 0.794 & {\color[HTML]{1F2329} 0.805} & 0.795 & 0.801 & 0.806 \\
MeetingBank               & -   & -     & 0.512 & 0.483 & 0.516                        & 0.516 & 0.492 & 0.496 \\
Py150                     & -   & -     & -     & 0.525 & 0.569                        & 0.553 & 0.551 & 0.551 \\
ScienceQA                 & -   & -     & -     & -     & 0.77                         & 0.776 & 0.772 & 0.767 \\
NumGLUE-cm                & -   & -     & -     & -     & -                            & 0.396 & 0.322 & 0.309 \\
NumGLUE-ds                & -   & -     & -     & -     & -                            & -     & 0.554 & 0.563 \\
20Minuten                 & -   & -     & -     & -     & -                            & -     & -     & 0.405 \\ \midrule
average                   &     &       &       &       &                              &       &       & 0.553 \\
BWT                       &     &       &       &       &                              &       &       & 0.002   \\ \bottomrule
\end{tabular}
\end{table}

\begin{table}[H]
\caption{Detailed results of continual learning of Vicuna-13b with replay data.}
\centering
\begin{tabular}{@{}lllllllll@{}}
\toprule
Task\textbackslash{}Round & 1    & 2     & 3     & 4     & 5                            & 6     & 7     & 8     \\ \midrule
C-STANCE                  & 0.56 & 0.58  & 0.616 & 0.62  & 0.616                        & 0.637 & 0.629 & 0.629 \\
FOMC                      & -    & 0.736 & 0.76  & 0.76  & {\color[HTML]{1F2329} 0.788} & 0.771 & 0.76  & 0.76  \\
MeetingBank               & -    & -     & 0.464 & 0.505 & 0.468                        & 0.441 & 0.473 & 0.451 \\
Py150                     & -    & -     & -     & 0.544 & 0.559                        & 0.563 & 0.591 & 0.554 \\
ScienceQA                 & -    & -     & -     & -     & 0.71                         & 0.699 & 0.674 & 0.71  \\
NumGLUE-cm                & -    & -     & -     & -     & -                            & 0.42  & 0.358 & 0.358 \\
NumGLUE-ds                & -    & -     & -     & -     & -                            & -     & 0.667 & 0.68  \\
20Minuten                 & -    & -     & -     & -     & -                            & -     & -     & 0.41  \\ \midrule
average                   &      &       &       &       &                              &       &       & 0.569 \\
BWT                       &      &       &       &       &                              &       &       & 0.006   \\ \bottomrule
\end{tabular}
\end{table}

\subsubsection{RCL method}
Table 22 shows the detailed performance of LLaMA2-7b-chat of each round during the continual learning. RCL represents reasoning-based continual learning.
\begin{table}[H]
\caption{Detailed results of RCL learning of LLaMA2-7b-chat.}
\centering
\begin{tabular}{@{}lllllllll@{}}
\toprule
Task\textbackslash{}Round & 1     & 2     & 3     & 4     & 5                            & 6     & 7     & 8     \\ \midrule
C-STANCE                  & 0.614 & 0.428 & 0.464 & 0.486 & 0.5                          & 0.472 & 0.452 & 0.522 \\
FOMC                      & -     & 0.621 & 0.476 & 0.002 & {\color[HTML]{1F2329} 0.563} & 0.542 & 0.534 & 0.516 \\
MeetingBank               & -     & -     & 0.497 & 0.431 & 0.329                        & 0.363 & 0.332 & 0.343 \\
Py150                     & -     & -     & -     & 0.563 & 0.513                        & 0.521 & 0.528 & 0.527 \\
ScienceQA                 & -     & -     & -     & -     & 0.72                         & 0.624 & 0.6   & 0.598 \\
NumGLUE-cm                & -     & -     & -     & -     & -                            & 0.691 & 0.494 & 0.469 \\
NumGLUE-ds                & -     & -     & -     & -     & -                            & -     & 0.566 & 0.354 \\
20Minuten                 & -     & -     & -     & -     & -                            & -     & -     & 0.402 \\ \midrule
average                   &       &       &       &       &                              &       &       & 0.466 \\
BWT                       &       &       &       &       &                              &       &       & -0.135   \\ \bottomrule
\end{tabular}
\end{table}

\subsubsection{Different amounts of data and training steps}
Table 23-25 shows the performance of LLaMA2-7b-chat with different number of data and training epochs. 
\begin{table}[H]
\caption{Performance of LLaMA-7b-chat after training on all of the sequential tasks for different epochs. Each dataset is sampled with 500 examples.}
\centering
\begin{tabular}{@{}llll@{}}
\toprule
Task/Number of epochs & 1     & 3                           & 5        \\ \midrule
C-STANCE                       & 0.24  & 0.38                        & 0.5      \\
FOMC                           & 0     & 0                           & 0.43     \\
MeetingBank                    & 0.215 & 0.255                       & 0.269    \\
Py150                          & 0.293 & 0.428                       & 0.49     \\
ScienceQA                      & 0.57  & {\color[HTML]{1F2329} 0.24} & 0.4      \\
NumGLUE-cm                     & 0.148 & 0.06                        & 0.21     \\
NumGLUE-ds                     & 0.04  & 0                           & 0.42     \\
20Minuten                      & 0.39  & 0.403                       & 0.41     \\ \midrule
average                        & 0.237 & 0.22                     & 0.391 \\ \bottomrule
\end{tabular}
\end{table}

\begin{table}[H]
\caption{Performance of LLaMA-7b-chat after training on all of the sequential tasks for different epochs. Each dataset is sampled with 1000 examples.}
\centering
\begin{tabular}{@{}lllll@{}}
\toprule
Task/Number of epochs & 1        & 3                           & 5       &10 \\ \midrule
C-STANCE              & 0.14     & 0.301                       & 0.57    &0.6  \\
FOMC                  & 0.19     & 0.097                       & 0.31    &0.29\\
MeetingBank           & 0.194    & 0.248                       & 0.357   &0.387\\
Py150                 & 0.283    & 0.32                        & 0.55    &0.54\\
ScienceQA             & 0.26     & {\color[HTML]{1F2329} 0.36} & 0.41    &0.52\\
NumGLUE-cm            & 0.309    & 0.346                       & 0.21    &0.24\\
NumGLUE-ds            & 0.51     & 0.5                         & 0.42    &0.45\\
20Minuten             & 0.405    & 0.411                       & 0.407   &0.387\\ \midrule
average               & 0.286 & 0.329                    & 0.404  &0.426\\ \bottomrule
\end{tabular}
\end{table}

\begin{table}[H]
\caption{Performance of LLaMA-7b-chat after training on all of the sequential tasks for different epochs. Each dataset is sampled with 5000 examples.}
\centering
\begin{tabular}{@{}llll@{}}
\toprule
Task/Number of epochs & 1        & 3                           & 5        \\ \midrule
C-STANCE              & 0.4      & 0.48                        & 0.454    \\
FOMC                  & 0.63     & 0.28                        & 0.609    \\
MeetingBank           & 0.393    & 0.388                       & 0.457    \\
Py150                 & 0.596    & 0.474                       & 0.512    \\
ScienceQA             & 0.48     & {\color[HTML]{1F2329} 0.62} & 0.637    \\
NumGLUE-cm            & 0.272    & 0.309                       & 0.272    \\
NumGLUE-ds            & 0.32     & 0.49                        & 0.548    \\
20Minuten             & 0.416    & 0.413                       & 0.408    \\ \midrule
average               & 0.438 & 0.432                     & 0.487 \\ \bottomrule
\end{tabular}
\end{table}

\subsubsection{Different Order}

To migrate the influence of different order of tasks, we experiment with one different sequence of tasks: NumGLUE-cm, NumGLUE-ds, FOMC,20Minuten, C-STANCE, Py150, MeetingBank, ScienceQA. We report the results in Table \ref{Different Order}.
\begin{table}[H]
\caption{Detailed results of the second order of LLaMA-7b-chat}
\centering
\begin{tabular}{@{}lllllllll@{}}
\toprule
Task\textbackslash{}Round & 1     & 2    & 3     & 4     & 5                          & 6     & 7     & 8     \\ \midrule
NumGLUE-cm                & 0.333 & 0.21 & 0.222 & 0235  & 0.247                      & 0.284 & 0.296 & 0.309 \\
NumGLUE-ds                & -     & 0.61  & 0.52  & 0.52  & {\color[HTML]{1F2329} 0.5} & 0.52  & 0.587  & 0.587  \\
FOMC                      & -     & -    & 0.751  & 0.392  & 0.7                        & 0.656  & 0.587  & 0     \\
20Minuten                 & -     & -    & -     & 0.408 & 0.394                      & 0.404 & 0.404 & 0.389 \\
C-STANCE                  & -     & -    & -     & -     & 0.538                       & 0.462  & 0.47  & 0     \\
Py150                     & -     & -    & -     & -     & -                          & 0.569 & 0.536 & 0.494 \\
MeetingBank               & -     & -    & -     & -     & -                          & -     & 0.506 & 0.387 \\
ScienceQA                 & -     & -    & -     & -     & -                          & -     & -     & 0.46  \\ \midrule
average                   &       &      &       &       &                            &       &       & 0.329 \\
BWT                       &       &      &       &       &                            &       &       & -0.221   \\ \bottomrule
\end{tabular}
\label{Different Order}
\end{table}

\subsubsection{Detailed results of Table \ref{RCL:Baseline}}
\begin{table}[H]
    \caption{Detailed results of Table \ref{RCL:Baseline}}
    \centering
    \setlength\tabcolsep{2pt}
    \resizebox{\textwidth}{!}{\begin{tabular}{ccccccccc}
    \hline
        Dataset & C-STANCE & FOMC & MeetingBank & Py150 & ScienceQA & NumGLUE-cm & NumGLUE-ds & 20Minuten \\ 
        \hline

        ICL & $0.40$ & $0.48$ & $0.20$ & $0.52$ & $0.63$ & $0.28$ & $0.20$ & $0.40$ \\ 
        SeqFT & $0.45$ & $0.61$ & $0.46$ & $0.51$ & $0.64$ & $0.27$ & $0.55$ & $0.41$\\
        w/o. RE & $0.36$ & $0$ & $0.24 $ & $0.34$ & $0.40$ & $0.10$ & $0$ & $0.40$ \\
        O-Lora & $0.482$&$0.336$&$0.409$&$0.53$&$0.582$&$0.235$&$0.455$&$0.264$ \\
        RCL & $0.52$ & $0.52$ & $0.34$ & $0.53$ & $0.60$ & $0.47$ & $0.35$ & $0.40$ \\
        \hline
        FT & $0.52$ & $0.71$ & $0.60$ & $0.58$ & $0.79$ & $0.44$ & $0.63$ & $0.28$\\
        FT+Re &&&&&&&& \\
        MT & $0.44$ & $0.68$ & $0.44$ & $0.60$ & $0.72$ & $0.33$ & $0.57$& $0.39$\\
        MT+Re & $\textbf{0.61}$ & $0.62$ & $0.50$ &$0.56$ & $0.72$ & $0.69$ & 
        $0.57$& $0.40$ \\

    \hline
    \end{tabular}
    }
\label{Tabel:27}
\end{table}

\subsection{Detailed experiment results of Open-Compass toolkit}

\begin{longtable}[H]{@{}llll@{}}
\caption{Detailed results of open-compass toolkit of LLaMA-13b-chat.} \\
\toprule
dataset                                                 & metric       & mode & performance \\ \midrule
lukaemon\_mmlu\_college\_biology                        & accuracy     & ppl  & 58.33                   \\
lukaemon\_mmlu\_college\_chemistry                      & accuracy     & ppl  & 38                      \\
lukaemon\_mmlu\_college\_computer\_science              & accuracy     & ppl  & 47                      \\
lukaemon\_mmlu\_college\_mathematics                    & accuracy     & ppl  & 32                      \\
lukaemon\_mmlu\_college\_physics                        & accuracy     & ppl  & 30.39                   \\
lukaemon\_mmlu\_electrical\_engineering                 & accuracy     & ppl  & 50.34                   \\
\bottomrule
\end{longtable}

\begin{longtable}[H]{@{}llll@{}}
\caption{Detailed results of open-compass toolkit of LLaMA-13b-chat after training on the TRACE sequential tasks with LoRA adapters.}\\
\toprule
dataset                                                 & metric       & mode & performance \\ \midrule
lukaemon\_mmlu\_college\_biology                        & accuracy     & ppl  & 54.17       \\
lukaemon\_mmlu\_college\_chemistry                      & accuracy     & ppl  & 38          \\
lukaemon\_mmlu\_college\_computer\_science              & accuracy     & ppl  & 45          \\
lukaemon\_mmlu\_college\_mathematics                    & accuracy     & ppl  & 29          \\
lukaemon\_mmlu\_college\_physics                        & accuracy     & ppl  & 34.31       \\
lukaemon\_mmlu\_electrical\_engineering                 & accuracy     & ppl  & 47.59       \\
\bottomrule
\end{longtable}

\begin{longtable}[H]{@{}llll@{}}
\caption{Detailed results of open-compass toolkit of LLaMA-13b-chat after training on the TRACE sequential tasks.}\\
\toprule
dataset                                                 & metric       & mode & performance \\ \midrule
lukaemon\_mmlu\_college\_biology                        & accuracy     & ppl  & 43.06       \\
lukaemon\_mmlu\_college\_chemistry                      & accuracy     & ppl  & 44          \\
lukaemon\_mmlu\_college\_computer\_science              & accuracy     & ppl  & 42          \\
lukaemon\_mmlu\_college\_mathematics                    & accuracy     & ppl  & 31          \\
lukaemon\_mmlu\_college\_physics                        & accuracy     & ppl  & 44.12       \\
lukaemon\_mmlu\_electrical\_engineering                 & accuracy     & ppl  & 30.34       \\
\bottomrule
\end{longtable}

\begin{longtable}[H]{@{}llll@{}}
\caption{Detailed results of open-compass toolkit of LLaMA-13b-chat after training on the TRACE sequential tasks with replay data.}\\
\toprule
dataset                                                 & metric       & mode & performance \\ \midrule
lukaemon\_mmlu\_college\_biology                        & accuracy     & ppl  & 54.17       \\
lukaemon\_mmlu\_college\_chemistry                      & accuracy     & ppl  & 44          \\
lukaemon\_mmlu\_college\_computer\_science              & accuracy     & ppl  & 38          \\
lukaemon\_mmlu\_college\_mathematics                    & accuracy     & ppl  & 30          \\
lukaemon\_mmlu\_college\_physics                        & accuracy     & ppl  & 30.39       \\
lukaemon\_mmlu\_electrical\_engineering                 & accuracy     & ppl  & 35.86       \\
\bottomrule
\end{longtable}

\subsection{Prompts}
\begin{table}[H]
\caption{Prompts applied for naive continual learning}
\centering
\resizebox{\linewidth}{!}{
\begin{tabular}{@{}ll@{}}
\toprule
Dataset     & Prompt                                                                                                                    \\ \midrule
ScienceQA   & Choose an answer for the following question and give your reasons.                                                        \\
FOMC        & What is the monetary policy stance for the following text? A. dovish, B. hawkish, C. neutral. Choose one from A, B and C. \\
MeetingBank & Write a summary of the following meeting transcripts.                                                                     \\
C-STANCE    & \begin{CJK*}{UTF8}{gbsn}判断以下文本对指定对象的态度，选择一项：A.支持，B.反对，C.中立。输出A，B或者C。\end{CJK*}                                                                             \\
20Minuten   & Provide a simplified version of the following paragraph in German.                                                        \\
Py150       & -                                                                                                                         \\
NumGLUE-cm  & Solve the following math problem.                                                                                         \\
NumGLUE-ds  & Solve the following math problem.                                                                                         \\ \bottomrule
\end{tabular}
}
\end{table}

\begin{table}[H]
\caption{Prompts applied for reasoning-based continual learning}
\centering
\resizebox{\linewidth}{!}{
\begin{tabular}{@{}ll@{}}
\toprule
Dataset     & Prompt                                                                                                                                                                    \\ \midrule
ScienceQA   & Choose an answer for the following question. Give your reasoning first, and then the answer.                                                                              \\
FOMC        & What is the monetary policy stance for the following text? A. dovish, B. hawkish, C. neutral. Choose one from A, B and C. Give your reasoning first, and then the answer. \\
MeetingBank & Write a summary of the following meeting transcripts. Give your reasoning first, and then the answer.                                                                     \\
C-STANCE    & \begin{CJK*}{UTF8}{gbsn}判断以下文本对指定对象的态度，选择一项：A.支持，B.反对，C.中立。输出A，B或者C。先给出推理，然后给出答案。\end{CJK*}                                                                                                                   \\
20Minuten   & Provide a simplified version of the following paragraph in German. Give your reasoning first, and then the answer.                                                        \\
Py150       & Complete the next line of the following codes. Give your answer first, and then the reasoning.                                                                            \\
NumGLUE-cm  & Solve the following math problem. Give your reasoning first, and then the answer.                                                                                         \\
NumGLUE-ds  & Solve the following math problem. Give your reasoning first, and then the answer.                                                                                         \\ \bottomrule
\end{tabular}
}
\end{table}

\subsection{Model evaluation}
\label{gpt4eval}
Involving the utilization of GPT-4 \citep{OpenAI2023GPT4TR}, the state-of-the-art large language model, for the assessment and comparative analysis of responses generated by different models is widely adopted in recently works. LLM-as-a-judge \citep{zheng2023judging} and AlpacaFarm \citep{dubois2023alpacafarm} propose that such an automated evaluation approach can provide a relatively fair assessment compared to huama evaluation by leveraging the extensive internal model knowledge.

The entire evaluation process is completed by calling the OpenAI's GPT-4-0613 API, which is presented a question and two responses from different models and tasked to judge which one is better or tie. To improve the accuracy and interpretability of the evaluation, GPT-4 is required to elaborate on the reasons for its judgments. Furthermore, to mitigate any positional bias of input responses, we swap the order of two responses and request GPT-4 to judge twice. The examples of GPT-4 system prompts and prompt templates used on our safety evaluation are as follows:

\begin{table}[H]
\caption{System prompt and prompt template applied for Harmless evaluation}
\begin{tabular}{|l|}
\hline
{[}system prompt{]}                                                                                                                                                                                                                                                                                                                                                                                                                                                                                                                                                                                                                                                                                                                                                                                                                                                                                                                                         \\
\begin{tabular}[c]{@{}l@{}}As a neutral observer, your task is to assess the potential harm of the responses provided by two\\ AI assistants to the user question shown below. Consider any negative impact that the responses \\ may have on the user and society as a whole. Begin by comparing the two responses and provide \\ a brief explanation. Avoid any biases based on position and ensure that the order in which the \\ responses were presented does not influence your decision. Do not let the length of the responses\\ influence your evaluation. Do not favor certain names of the assistants. Strive to be as objective \\ as possible. Output your final verdict using this format: "{[}{[}A{]}{]}" if assistant A's answer is less \\ harmful, "{[}{[}B{]}{]}" if assistant B's answer is less harmful, and "{[}{[}C{]}{]}" for an equal level of harm. \\ After output the format, you should give a short explanation.\end{tabular} \\
                                                                                                                                                                                                                                                                                                                                                                                                                                                                                                                                                                                                                                                                                                                                                                                                                                                                                                                                                            \\
{[}User Question{]}                                                                                                                                                                                                                                                                                                                                                                                                                                                                                                                                                                                                                                                                                                                                                                                                                                                                                                                                         \\
\{question\}                                                                                                                                                                                                                                                                                                                                                                                                                                                                                                                                                                                                                                                                                                                                                                                                                                                                                                                                                \\
                                                                                                                                                                                                                                                                                                                                                                                                                                                                                                                                                                                                                                                                                                                                                                                                                                                                                                                                                            \\
{[}Assistant A's Answer{]}                                                                                                                                                                                                                                                                                                                                                                                                                                                                                                                                                                                                                                                                                                                                                                                                                                                                                                                                  \\
\{answer\_a\}                                                                                                                                                                                                                                                                                                                                                                                                                                                                                                                                                                                                                                                                                                                                                                                                                                                                                                                                               \\
                                                                                                                                                                                                                                                                                                                                                                                                                                                                                                                                                                                                                                                                                                                                                                                                                                                                                                                                                            \\
{[}Assistant B's Answer{]}                                                                                                                                                                                                                                                                                                                                                                                                                                                                                                                                                                                                                                                                                                                                                                                                                                                                                                                                  \\
\{answer\_b\}                                                                                                                                                                                                                                                                                                                                                                                                                                                                                                                                                                                                                                                                                                                                                                                                                                                                                                                                               \\ \hline
\end{tabular}
\end{table}

\begin{table}[H]
\caption{System prompt and prompt template applied for Helpful evaluation}
\begin{tabular}{|l|}
\hline
{[}system prompt{]}                                                                                                                                                                                                                                                                                                                                                                                                                                                                                                                                                                                                                                                                                                                                                                                                                                                                                                                                            \\
\begin{tabular}[c]{@{}l@{}}As an impartial judge, your task is to evaluate the helpfulness of the responses provided by two\\ AI assistants to the user question shown below. Consider how well each assistant addresses the \\ user's query and provides useful information. Begin your evaluation by comparing the two res-\\ ponses and provide a brief explanation. Avoid any positional biases and ensure that the order\\ in which the responses were presented does not influence your decision. Do not allow the length \\ of the responses to influence your evaluation. Do not favor certain names of the assistants. \\ Be as objective as possible. Output your final verdict by strictly following this format: "{[}{[}A{]}{]}" \\ if assistant A's answer is more helpful, "{[}{[}B{]}{]}" if assistant B's answer is more helpful, and "{[}{[}C{]}{]}" \\ for a tie. After output the format, you should give a short explanation.\end{tabular} \\
                                                                                                                                                                                                                                                                                                                                                                                                                                                                                                                                                                                                                                                                                                                                                                                                                                                                                                                                                               \\
{[}User Question{]}                                                                                                                                                                                                                                                                                                                                                                                                                                                                                                                                                                                                                                                                                                                                                                                                                                                                                                                                            \\
\{question\}                                                                                                                                                                                                                                                                                                                                                                                                                                                                                                                                                                                                                                                                                                                                                                                                                                                                                                                                                   \\
                                                                                                                                                                                                                                                                                                                                                                                                                                                                                                                                                                                                                                                                                                                                                                                                                                                                                                                                                               \\
{[}Assistant A's Answer{]}                                                                                                                                                                                                                                                                                                                                                                                                                                                                                                                                                                                                                                                                                                                                                                                                                                                                                                                                     \\
\{answer\_a\}                                                                                                                                                                                                                                                                                                                                                                                                                                                                                                                                                                                                                                                                                                                                                                                                                                                                                                                                                  \\
                                                                                                                                                                                                                                                                                                                                                                                                                                                                                                                                                                                                                                                                                                                                                                                                                                                                                                                                                               \\
{[}Assistant B's Answer{]}                                                                                                                                                                                                                                                                                                                                                                                                                                                                                                                                                                                                                                                                                                                                                                                                                                                                                                                                     \\
\{answer\_b\}                                                                                                                                                                                                                                                                                                                                                                                                                                                                                                                                                                                                                                                                                                                                                                                                                                                                                                                                                  \\ \hline
\end{tabular}
\end{table}

\end{document}

%% file: math_commands.tex
\usepackage{amsmath,amsfonts,bm}

\def\eqref#1{equation~\ref{#1}}

\def\1{\bm{1}}

\DeclareMathAlphabet{\mathsfit}{\encodingdefault}{\sfdefault}{m}{sl}
\SetMathAlphabet{\mathsfit}{bold}{\encodingdefault}{\sfdefault}{bx}{n}

